\documentclass[10pt,twocolumn,letterpaper]{article}

\usepackage[pagenumbers]{cvpr} 

\newcommand{\graph}{\mathcal{G}}

\newcommand{\edge}{E}

\newcommand{\rel}{T}

\DeclareMathOperator*{\argmin}{arg\,min}

\newcommand{\icon}[2][1.2em]{%
  \raisebox{-0.15\height}{%
    \includegraphics[height=#1]{#2}%
  }%
}

\newcommand{\gain}[1]{#1}

\newif\ifshowcomments
\showcommentstrue      

\definecolor{tableblue}{RGB}{209,226,252}
\definecolor{lighttableblue}{RGB}{244,249,254}
\definecolor{tablelightyellow}{RGB}{255,207,145}
\definecolor{tableyellow}{RGB}{255,239,213}

\newcommand\blfootnote[1]{%
  \begingroup
 \renewcommand\thefootnote{}\footnote{#1}%
  \addtocounter{footnote}{-1}%
  \endgroup
}

\definecolor{cvprblue}{rgb}{0.21,0.49,0.74}
\usepackage[pagebackref,breaklinks,colorlinks,allcolors=cvprblue]{hyperref}
\usepackage{multirow}
\usepackage{amsmath}
\usepackage{makecell}
\usepackage{gensymb}
\usepackage{booktabs}
\usepackage[export]{adjustbox}
\usepackage{fontawesome6}
\usepackage{pifont}
\usepackage[table]{xcolor}

\def\paperID{58} 
\def\confName{3DV\xspace}
\def\confYear{2027\xspace}

\title{Fast Pose Tracking of Rigid Objects\\ with Compact Pose Graph Optimization}

\author{Xiaojie Zhang$^{1}$
\quad
Tom Fischer$^{1}$
\quad
Viktor Larsson$^{2}$ \quad
\quad
Eddy Ilg$^{1\dagger}$\\
\\
$^{1}$University of Technology Nuremberg \quad $^{2}$Lund University
}

\begin{document}

\maketitle
\blfootnote{$\dagger$ Now at Google}

\begin{abstract}
Tracking a novel object's 6D pose over long horizons currently requires either expensive onboarding or a reconstruction maintained throughout the sequence. 
This makes current trackers impractical for robotic manipulation and augmented reality, which need trackers that are ready to use and run in real time.
We show that a lightweight tracking module can be applied on top of a wide range of correspondence estimation methods to keep drifts bounded while maintaining fast runtime.
Our key idea is to avoid point-based optimization in the pose graph and operate only on relative pose constraints, which we weight by a derived uncertainty from the geometric alignment. 
This makes optimization independent of the number of correspondences while avoiding the direct inclusion of noisy point measurements, leading to fast and robust long-term tracking.
Across four real-world benchmarks, our approach achieves tracking accuracy comparable to reconstruction-based trackers with a fraction of the optimization cost. Overall, these results suggest that a compact and reliable pose graph optimization can provide long-horizon consistency at substantially lower computational cost. 

\end{abstract}

\section{Introduction}
\label{sec:intro}

Tracking an object's 6D pose over time is important for understanding dynamic environments, with applications in robotic manipulation~\cite{robotmanip1,
surveyTrack}, augmented reality~\cite{surveyAR}, and human--object
interaction~\cite{surveyTrack2}. 
Two requirements make this problem challenging in practice: the tracking must run online at interactive rates to support downstream planning, and it must remain accurate and temporally consistent over long horizons despite occlusions and fast motions.

Existing approaches satisfy these two requirements one at a time. 
Long-horizon accuracy is typically obtained by anchoring poses to a 3D representation, either built online during tracking~\cite{bundlesdf,6dopegs,dengmonotracker} or onboarded beforehand from CAD models or posed reference frames~\cite{foundationpose}.
While both approaches can be accurate, the first is too slow for real-time use and the second must be repeated for every new object. 
Interactive rates without onboarding can be achieved by relative pose estimation methods~\cite{one2any,unopose,any6d}, which estimate pose from a single reference observation.
However, these operate only on pairs of frames, so small errors in relative motion compound over time and cause drift.

We show that both requirements can be met by constraining pairwise measurements against each other in a pose graph without maintaining a 3D representation.
While pose graph optimization (PGO) is widely used in vision and robotics, existing object trackers typically build their graph constraints directly from correspondences or dense geometric observations. 
We instead ask: once relative motion has been estimated from a set of correspondences, is it necessary to retain those points in the global optimization?

Our key idea is to compress each set of correspondences into a relative pose measurement and perform global optimization only over poses.
This has two advantages. 
When correspondences are reliable, many inliers may be available and the relative pose is already well constrained. Retaining all of those points in the global objective only increases the optimization cost, whereas our update cost is independent of the number of inliers.
When correspondences are poor, directly placing noisy points in the global objective also exposes the optimization to their errors. 
By operating only on relative pose constraints, we avoid both costs.

\begin{figure*}[t]
    \centering
    \includegraphics[width=\linewidth]{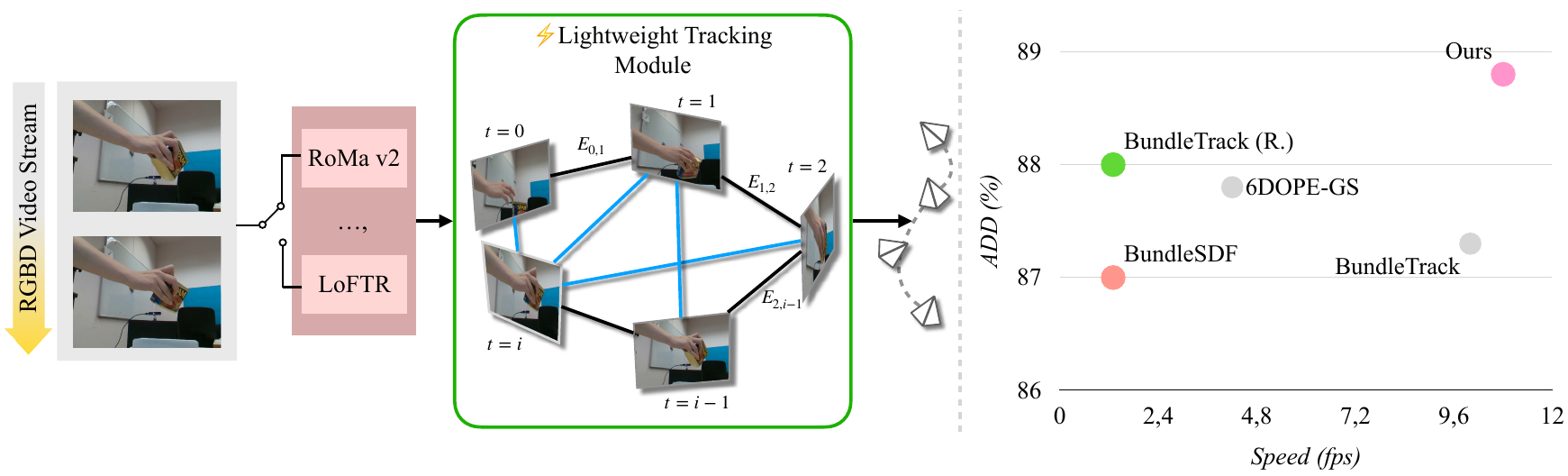}
    \caption{\textbf{Left:} We propose a lightweight generalizable 6D pose tracking framework for novel objects that can be flexibly paired with any correspondence estimation method. \textbf{Right:} We show that our approach achieves the best pose accuracy and runtime efficiency tradeoff on the YCBInEOAT dataset. Gray: reported in the original paper on different hardware, see \cref{tab:run_diff_gpus}.
    }
    \label{fig:teaser}
\end{figure*}

This formulation shifts the main question from how many correspondences are retained to how much each relative pose measurement should be trusted. 
We therefore derive a pose uncertainty from the geometric alignment that produced each relative transform and use it to weight the corresponding graph constraint. 
Poorly constrained measurements contribute less, while reliable measurements are enforced more strongly. 
The result is a lightweight tracking module that can be paired with a wide range of correspondence estimation methods (\cref{fig:teaser}) and, in our experiments, achieves accuracy comparable to reconstruction- and CAD-based methods at substantially lower runtime.

In summary, our contributions are as follows:

\begin{itemize}
    \item
     We formulate 6D object tracking as a compact pose graph optimization over relative pose constraints. By compressing dense correspondences into an $SE(3)$ measurement, the optimization cost is independent of the number of matched correspondences, enabling substantially faster updates while retaining global trajectory consistency.

    \item We derive per-edge pose uncertainty from the geometric alignment and use it for weighting pose graph constraints. This allows poorly constrained relative poses to contribute less to the trajectory estimation and improves robustness when the correspondence quality degrades.

    \item The proposed tracker can be paired with different correspondence or relative-pose estimators without retraining or per-object onboarding. Across four real-world RGB-D benchmarks, our method achieves accuracy comparable to state-of-the-art tracking-by-reconstruction approaches across four benchmarks with a fraction of the optimization cost.
    
\end{itemize}

\section{Related Work}
\label{sec:related_work}
\subsection{Pose Estimation}
Estimating camera poses is central for scene- and object-level understanding.
A common approach is to estimate correspondences through feature matching and then solve camera poses through classical geometry, such as the eight-point algorithm~\cite{hartley1997defense} or Perspective-n-Point (PnP)~\cite{epnp}.
Scene-level pipelines~\cite{reposed,colmap} typically focus on relative transformations between RGB images and rely on (dense) 2D--2D correspondences.
To improve robustness across diverse scenes and large viewpoint changes, recent approaches adopt deep learning-based matchers~\cite{lofter,roma,romav2,lightglue,superglue} trained on large-scale data, which substantially outperform hand-crafted features~\cite{sift,orb}.

In object-centric settings, pose estimation methods~\cite{foundationpose, gen6d, secondpose,fischer2025unified,sundermeyer2018implicit} often learn 3D object representations to model geometry and appearance. 
Relative pose estimation methods for novel objects~\cite{unopose,one2any,any6d} instead predict the relative transformation between two frames without requiring instance-specific object knowledge (e.g. CAD models, multiple reference frames). 
These methods can generalize to unseen objects and can handle moderate viewpoint changes.
In this work, we propose a general tracking module which can be paired with both 2D feature matchers and relative object pose estimation methods and improves temporal consistency by fusing pairwise measurements over time.

\subsection{Pose Graph Optimization}

Pose graph optimization (PGO)~\cite{carlone2012linear,carlone2018convex} estimates absolute poses from pairwise constraints by enforcing consistency over a graph.
It has been widely used across structure-from-motion~\cite{pan2024global,zhang2023revisiting}, SLAM~\cite{maggio2026vggt}, and robotics~\cite{factor_graphs_for_robot_perception}. 
Importantly, different systems instantiate PGO with different residual models, where some optimize alignment-based residuals defined directly on correspondences~\cite{bundlefusion}, while others represent pairwise relations as relative-pose factors and optimize pose-consistency residuals in $SE(3)$ poses~\cite{gtsam,zhang2023revisiting,pan2024global}.
While both types are common in scene-level pipelines, pose-consistency terms have received comparatively little attention in object-centric RGB-D tracking, where prior work more commonly relies on alignment-based objectives. Dellaert \textit{et al.} introduced factor-graph formulations and the \textit{Bayes tree} data structure for incremental smoothing and mapping (iSAM2)~\cite{isam2}, which supports low-latency updates without periodic batch steps through fluid relinearization and incremental variable reordering.
In this work, we integrate iSAM2 to efficiently update our 6D object pose graph online as new frames arrive.

\subsection{Object 6D Pose Tracking}

BundleTrack~\cite{bundletrack} is an early framework for novel-object 6D pose tracking that does not require CAD models.
It combines feature matching with a memory-augmented pose graph that is optimized through an alignment loss on unprojected point clouds (similar to~\cite{bundlefusion}).
BundleSDF~\cite{bundlesdf} extends this line of work by integrating an online object reconstruction into the tracking loop, which can improve long-horizon stability but further increases computational cost.
To reduce optimization overhead, 6DOPE-GS~\cite{6dopegs} proposes a 2D Gaussian-Splatting-based representation~\cite{2dgs} for faster online reconstruction and reports a speed--accuracy trade-off.

Overall, alignment-based objectives in prior object-centric pose-graph trackers~\cite{bundletrack,bundlesdf,6dopegs} can yield accurate pose estimates. However, they couple the dense correspondence inside the global optimization, making optimization cost scale with the number of points or image resolutions. Additionally, this makes them more prone to error on data with noisy depth and degraded correspondence quality. In contrast, we use correspondences only to generate relative pose measurements and then optimize a compact pose graph over relative pose constraints. This design preserves modularity with respect to the correspondence estimator while enabling efficient and robust online updates.
\section{Methodology}

\begin{figure*}[t]
    \centering
    \includegraphics[width=\linewidth]{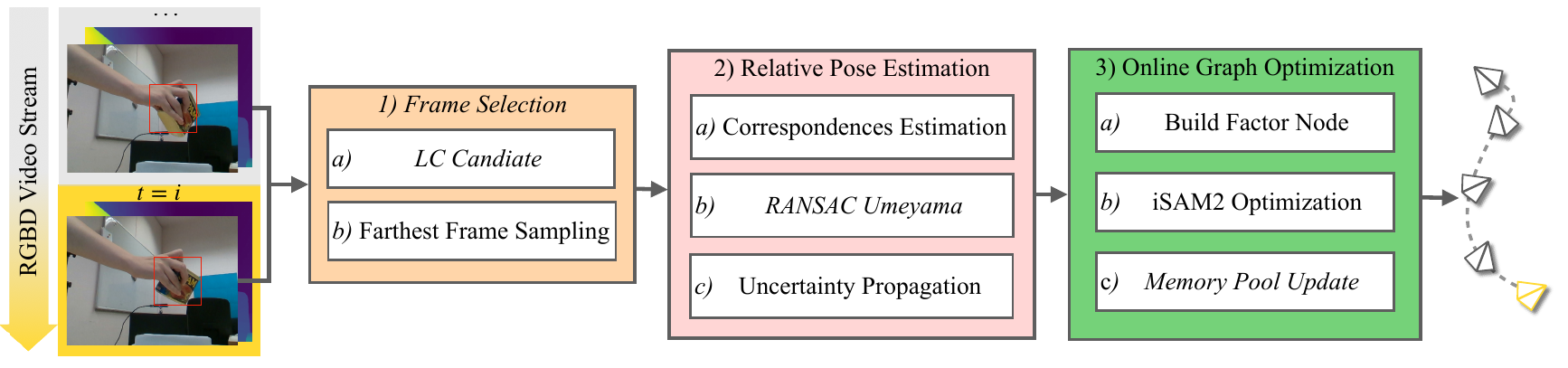}
    \caption{\textbf{High-level method overview.} We initialize a $SE(3)$ pose graph which will be incrementally updated when a new frame arrives. In the first stage of tracking (\textit{1}), we select frames from the past which we want to connect to the current frame via edges in the pose graph. Next, for each pair of frames corresponding to an edge, we estimate their correspondences (\textit{2.a}), and use RANSAC Umeyama to solve 6D relative poses (\textit{2.b}) and estimate their uncertainty(\textit{2.c}). Finally, we insert (\textit{3.a}) and optimize (\textit{3.b}) the new node and its edges to the pose graph using iSAM2 , and update the memory pool (\textit{3.c}).   
    }
    \label{fig:pipeline}
\end{figure*}

We track the 6D pose of an object across an RGB-D video by maintaining a compact $SE(3)$ pose graph that is extended and optimized incrementally (\cref{fig:pipeline}) over relative pose constraints. Every new frame contributes an odometry constraint to its predecessor and up to $K$ loop closure constraints to past keyframes, where each is weighted by an uncertainty derived from the conditioning of the geometric alignment problem.
The formulation is agnostic to how correspondences are obtained and can turn any feature matcher into a 6D object pose tracker.

\subsection{Problem Formulation}
Given an RGB-D video $\{(I_i, D_i)\}_{i=0}^{N}$ with images, depth maps, and masks of the target object $\{M_i\}_{i=0}^{N}$, our task is to track the object's 6D pose $T_i\in SE(3)$, where $T_i$ denotes the pose of frame $I_i$ relative to $I_0$.
We assume that the whole tracking process is online such that at time step $i$ a tracker only has access to the past frames $\{0,\dots,i\}$.

\subsection{Object Pose Graph Optimization}

We frame object pose tracking as optimization over a pose graph $\graph=(\mathcal{V},\mathcal{E})$, in which each node $V_i\in\mathcal{V}$ is the pose $T_i$ of frame ${I}_i$ and each edge  $\edge_{ij}\in\mathcal{E}$ carries a relative measurement $\rel_{ij}$ between the two frames. The residual $L_e\in\mathbb{R}^6$ of edge $\edge_{ij}$ is
\begin{equation}
\label{eq:pose_res}
L_{e}(T_i,T_j) = \mathrm{Log}\!\left( \rel_{ij}^{-1}\, (T_i^{-1}T_j) \right),
\end{equation}
where $\mathrm{Log}: SE(3)\rightarrow\mathbb{R}^6$ is the logarithm map to the tangent space. We anchor the first node at $T_0=\mathbb{I}$ to fix the gauge freedom, and then solve
\begin{align}
\label{eq:weighted_pgo}
\{T_i^*\}_{i=0}^N = \argmin_{\{T_i\}_{i=0}^N} \sum_{(i,j)\in\mathcal{E}} \rho\!\left( \left\| L_{e}(T_i,T_j)\right\|_{\Sigma_{ij}} \right),
\end{align}
where $\rho(\cdot)$ is a robust loss, $\|e\|_{\Sigma}^{2}=e^{\top}\Sigma^{-1}e$ denotes the Mahalanobis norm, and $\Sigma_{ij}$ is the uncertainty of the measurement $\rel_{ij}$.

\subsection{Frame Selection}
\label{sec:edge selection}

Choosing reliable keyframes is crucial to avoid drift that would accumulate when chaining odometries alone without long-range constraints through loop closure edges.
Deciding which past frames to connect to a new frame therefore determines how much of edges are informative.

When a new frame $I_i$ arrives, we select up to $K$ candidates for loop closures from a memory pool $\mathcal{M}$ containing past keyframes.
To avoid estimating $\rel_{i-1,i}$ before selection, we extrapolate a motion prior $T_i'$ under a constant-velocity model,
\begin{equation}
\label{eq:motion_prior}
    T_i' = T_{i-1}\,(T_{i-2}^{-1}T_{i-1}),
\end{equation}
such that we only require a single forward pass for the correspondences, and $T_i'$ is only used to guide frame selection. We then retain the frames in $\mathcal{M}$ whose rotation lies within radius $\tau_k$ of the rotation component of $T_i'$:
\begin{equation}
\label{eq:frame_selection}
    d_{geo}(R^{'}_i, R_j) = \arccos\!\left(\frac{\mathrm{tr}(R_j^{\top}R_i^{'})-1}{2}\right) \leq \tau_k .
\end{equation}
We select candidates from similar rotations, since we found that similar poses generally have more visual overlap, yield more matches and more correct relative pose estimation. From these candidates we sample a final set $\mathcal{K}$ of at most $K$ frames via farthest-frame sampling over timesteps, analogous to farthest-point sampling for point clouds. This results in edges of varying lengths that maximize the information gain. Shorter edges retain larger overlap and
are easier to match, whereas longer edges anchor the current frame to distinct parts of the trajectory rather than reinforcing what odometry already constrains.

\subsection{Relative Pose Estimation}
\textbf{Pose from Correspondences.}
\label{sec:relative pose estimation}
We estimate the relative poses between $I_i$ and the keyframes $\mathcal{K}$ from any source of correspondences, e.g. 2D-2D~\cite{roma, romav2, lofter}, 3D-3D~\cite{unopose,one2any}. 
For 2D-2D matchers, we backproject the correspondences with depth.
In both cases we obtain a set of correspondences
\begin{equation}
\mathcal{C}^{\text{full}}_{ik}=\{(\mathbf{a}_m,\mathbf{b}_m,w_m)\}_{m=1}^{M},
\end{equation}
where $\mathbf{a}_m,\mathbf{b}_m\in\mathbb{R}^3$ are corresponding 3D points in $I_k$ and $I_i$, and $w_m$ are confidence weights output by the feature matcher.
We run RANSAC to identify the inlier subset $\mathcal{C}_{ik} \subseteq \mathcal{C}^{\text{full}}_{ik}$, and solve for $\rel_{ik}$ via weighted Umeyama over inliers,
\begin{equation}
\label{eq:wop}
(R_{ik},t_{ik})
= \argmin_{R,\,t}
\sum_{(\mathbf{a},\mathbf{b},w)\,\in\,\mathcal{C}_{ik}} w \,\left\| R\,\mathbf{b} + t - \mathbf{a} \right\|_2^2 ,
\end{equation}
which admits a closed-form solution via weighted SVD~\cite{umeyama1991least}.

\noindent\textbf{Uncertainty Propagation.}
\label{sec:uncertainty propagation}
Common failure modes in object pose estimation, such as thin objects or heavy occlusion lead to poor conditioning of the alignment due to noisy or few correspondences.
Since this is a property of the alignment, we can measure it and downweight corresponding edges in the pose graph. We define a measurement noise $\eta_{ik} \in \mathbb{R}^{6}$ of the relative pose $\rel_{ik}$ in the tangent space:
\begin{align}
    T_i = T_k\rel_{ik}\text{Exp}(\eta), \space \eta \sim \mathcal{N}\!\left(\mathbf{0}_{6\times 1},\, \Sigma_{ik}\right) \mathrm{.}
\end{align}
We model the uncertainty as the covariance $\Sigma_{ik}$ of the pose estimated by~\cref{eq:wop} which we approximate with the inverse Hessian, following~\cite{triggs1999bundle}:
\begin{equation}
\label{eq:gn_info}
H_{ik} \approx \sum_{m=1}^{|\mathcal{C}_{ik}|} J_m^\top J_m, 
\qquad
\Sigma_{ik} \approx \sigma_{ik}^2 H_{ik}^{-1},
\end{equation}
where $J$ is the Jacobian of the whitened residual $\tilde{r}_m = \sqrt{w}\,r$ and $r_m = R_{ik}\mathbf{b}_m + t_{ik} - \mathbf{a}_m$, $\sigma_{ik}^2$ is the estimated residual variance to calibrate the scale of $\Sigma$. Without $\sigma_{ik}$, the absolute scale of $\Sigma_{ik}$ is set by an arbitrary scale of the confidences of the matchers. We parameterize the pose increment in the tangent space by $\delta\xi = (\delta\theta,\delta t)\in\mathbb{R}^6$ and use a right-multiplicative rotation perturbation $ R_{ik}\text{Exp}([\delta\theta]_\times)$. Linearizing $r_m$ yields
\begin{align}
\tilde{r}_m(\delta\xi) \approx \tilde{r}_m(0) + J_m\,\delta\xi
, J_m =
\sqrt{w_m}\,
\begin{bmatrix}
- R_{ik}[\mathbf{b}_m]_\times & \mathbf{I}
\end{bmatrix},
\label{eq:jac}
\end{align}
where $[\cdot]_\times$ denotes the skew-symmetric matrix, $\mathbf{I}$ is a $3\times3$ identity matrix. Stacking $J_m$ over correspondences leads to the full Jacobian defined in~\cref{eq:gn_info}. The full derivation is provided in the \emph{supplementary material}. 

\begin{figure*}[t]
    \centering
    \includegraphics[width=\linewidth]{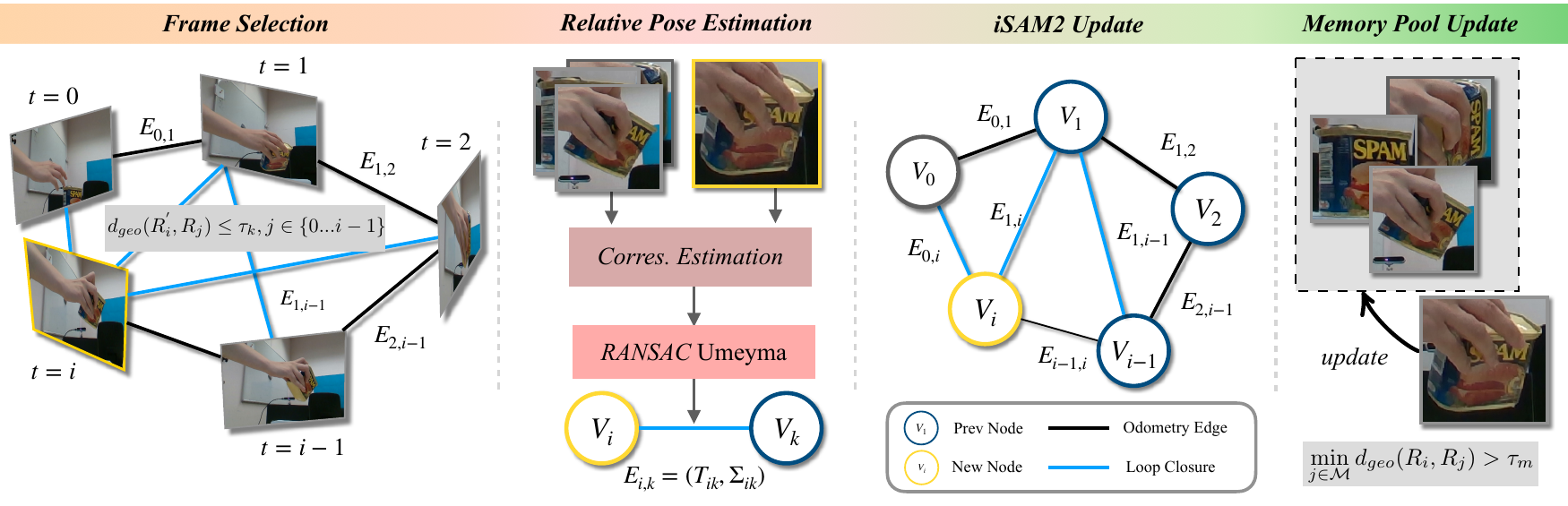}
    \caption{\textbf{Detailed method overview.} For a new frame $I_i$, we first select $K$ frames from previous frames to build candidate loop closure constraints based on the geodesic distance and farthest frame sampling. Then the current frame $I_i$ and selected $K$ candidate frames will be passed to the correspondence estimation method and RANSAC Umeyama to obtain relative poses $T_{ik}$ and uncertainty $\Sigma_{ik}$. Furthermore, we insert the new node $V_i$ with its odometry and loop closure edges to the uncertainty-aware graph and update $T_i$ using iSAM2. Last, we update the memory pool $\mathcal{M}$ based on viewpoint diversity. }
    \label{fig:pipeline2}
\end{figure*}

\subsection{Online Pose Graph Optimization}
\label{sec:online PGO}

\noindent\textbf{iSAM2 Update.}
We insert the new node $V_i$, its odometry edge $E_{i-1.i}$, and loop closure constraints to the pose graph $\graph$. 
We update the pose graph using iSAM2~\cite{isam2}. 
Internally, iSAM2 maintains a sparse factorization of the linearized system using a Bayes-tree representation and performs selective relinearization, so that only variables affected by new constraints are revisited. 
This yields low-latency updates suitable for online 6D tracking.
For our online setting, previously estimated poses will also be optimized but only for optimization of future frames. 

\noindent\textbf{Memory Pool Update.}
\label{sec:mem_pool_update}
Finally, we update the memory pool $\mathcal{M}$ from which loop closure candidates are
drawn. 
Its purpose is to keep the candidate set viewpoint-diverse, so that the
farthest-frame sampling in \cref{sec:edge selection} operates over distinct views rather than near-duplicates.
The pool $\mathcal{M}$ is initialized with the first frame and extended by the current frame $I_i$ if its geodesic distance to \emph{every} frame in $\mathcal{M}$ exceeds a threshold $\tau_m$, i.e., $\min_{j \in \mathcal{M}} d_{geo}(R_i, R_j) > \tau_m$.

\section{Evaluation}
In this section, our experiments address four main questions: (1) How does our model compare to state-of-the-art methods in both pose tracking accuracy and runtime efficiency? (2) How sensitive is our method to the quality of correspondences? (3) Does the resulting tracker remain stable over long sequences? (4) How important are the different components of our model?
\subsection{Setup}
\textbf{Metrics.}
We follow the evaluation protocol of~\cite{bundlesdf,bundletrack,foundationpose} and evaluate 6D pose tracking quality using the area under the curve (AUC) percentage for ADD-S and ADD~\cite{posecnn} ($0\sim0.1m$) using ground-truth object models. These metrics measure the mean distance between corresponding points on the predicted and ground-truth object models, providing a robust measure of geometric alignment accuracy. Inference speed is reported in frames per second (fps).\\
\textbf{Benchmarks.}
We evaluate our method on four real-world object pose tracking benchmarks, HOI4D~\cite{hoi4d}, YCBInEOAT~\cite{wen2020se}, HO3D~\cite{ho3d} and YCB Video~\cite{posecnn}. 
\textbf{HOI4D}~\cite{hoi4d} is a large-scale egocentric hand-object interaction dataset, containing 973 videos for 49 rigid objects from 5 categories (toy car, kettle, knife, bowl, chair). The average video length is 300 frames. \textbf{YCBInEOAT}~\cite{wen2020se} is an ego-centric robot manipulation dataset that consists of 9 scenes and 5 objects, with 823 frames on average.
\textbf{HO3D}~\cite{ho3d} is a hand-object interaction dataset that contains 13 scenes and 4 objects, with average 1547 frames. \textbf{YCB-Video}~\cite{posecnn} contains 21 objects in 12 cluttered scenes with occlusions and symmetric objects. The average length is 1718 frames.

\noindent\textbf{Baselines.}
We compare our method with other novel-object 6D pose tracking methods and divide them into four categories: {1)} Tracking-by-Reconstruction (\textbf{TbR}): BundleSDF~\cite{bundlesdf} and 6DOPE-GS~\cite{6dopegs}; 
{2)} Relative object Pose Estimation (\textbf{RPE}): UNOPose~\cite{unopose} and One2Any~\cite{one2any}; 
{3)} Pose Graph Optimization-based (\textbf{PGO}): BundleTrack~\cite{bundletrack}; {4) CAD-based: FoundationPose~\cite{foundationpose}}.  
We note that 4) requires either \emph{known} object models or reconstruction from posed reference views and is therefore not directly comparable.

\subsection{Implementation}
For our main method, we use RoMa v2~\cite{romav2} (turbo) for feature matching. Unless otherwise specified, we use $K=5$ as the default number.
For online graph optimization,  we use iSAM2 from the GTSAM~\cite{gtsam} library. 
We evaluate causal poses output at the time of arrival w.o. iSAM2 smoothing. 
To evaluate against ground-truth pose annotations, we align the predicted trajectory by applying the relative transform between our first-frame prediction and the ground-truth pose of the first frame. 
For 6DOPE-GS~\cite{6dopegs}, we report their results from the original paper due to its code currently not being available. 
All experiments and baselines were conducted on an NVIDIA H100 GPU. We use identical masks for all methods.

\subsection{Comparison to State-of-the-Art}
\begin{table*}[t]
\centering
\resizebox{\linewidth}{!}{%
\begin{tabular}{l l cc ccc ccc ccc c}
\toprule[1.2pt]
\multirow{2}{*}{\textbf{Cat.}} 
& \multirow{2}{*}{\textbf{Method}}
& \multicolumn{2}{c}{\makecell{\textbf{HOI4D}\\(300 frames)}}
& \multicolumn{3}{c}{\makecell{\textbf{YCBInEOAT}\\(823 frames)}}
& \multicolumn{3}{c}{\makecell{\textbf{HO3D}\\(1547 frames)}}
& \multicolumn{3}{c}{\makecell{\textbf{YCB Video}\\(1718 frames)}}
& \multirow{2}{*}{\makecell{\textbf{Avg. Speed}\\\textbf{fps}}}\\
\cmidrule(lr){3-4}
\cmidrule(lr){5-7}
\cmidrule(lr){8-10}
\cmidrule(lr){11-13}
& & {\scriptsize ADD-S (\%)} & {\scriptsize ADD (\%)}
& {\scriptsize ADD-S (\%)} & {\scriptsize ADD (\%)} & {\scriptsize AR (\%)}
& {\scriptsize ADD-S (\%)} & {\scriptsize ADD (\%)} & {\scriptsize AR (\%)}
& {\scriptsize ADD-S (\%)} & {\scriptsize ADD (\%)} & {\scriptsize AR (\%)} \\
\midrule

\textbf{CAD} & FoundationPose (CAD)
& 84.4 & 66.5
& 96.4 & 93.1 & 90.5
& 97.0 & 88.9 & 89.3
& 98.0 & 96.0 & 92.8
& 22.9\\
\midrule
\multirow{2}{*}{\textbf{RPE}}
& UNOPose
& \cellcolor{tableblue}\textbf{92.2} & 79.4
& 91.0 & 67.7 & 65.7
& 92.9 & 51.6 & 31.7
& 96.4 & 84.7 & 80.6
& 10.0 \\
& One2Any
& 79.4 & 64.0 
& 66.0 & 31.5 & 31.1
& 75.5 & 37.1 & 29.4
& 92.9 & 88.7 & 75.0
& 30.0 \\
\cmidrule{2-13}
\multirow{2}{*}{\textbf{TbR}}
& BundleSDF
& 90.8 & 79.0
& \cellcolor{lighttableblue}{93.8} & 87.0 & 82.2
& \cellcolor{tableblue}\textbf{96.5} & \cellcolor{tableblue}\textbf{92.6} & \cellcolor{lighttableblue}{87.9}
& \cellcolor{lighttableblue}{97.6} & 93.8 & 90.0
& 1.8 \\
& 6DOPE-GS$\dagger$
& - & -
& \cellcolor{lighttableblue}{93.8} & 87.8 & -
& 95.1 & 84.3 & -
& - & - & -
& $4.2$ \\
\cmidrule{2-13}
\multirow{3}{*}{\textbf{PGO}}
& BundleTrack$\dagger$
& - & -
& 92.5 & 87.3 & 79.0
& 92.4 & 66.0 & -
& - & - & -
& $10.0$ \\
& BundleTrack (R.)
& \cellcolor{lighttableblue}91.8 & \cellcolor{lighttableblue}{79.8}
& \cellcolor{lighttableblue}{93.8} & \cellcolor{lighttableblue}{88.0} & \cellcolor{lighttableblue}{82.7}
& \cellcolor{tableblue}\textbf{96.5} & \cellcolor{tableblue}\textbf{92.6} & \cellcolor{tableblue}\textbf{88.0}
& 97.5 & \cellcolor{lighttableblue}{94.9} & \cellcolor{lighttableblue}{91.8}
& 1.4 \\
& Ours
& \cellcolor{tableblue}\textbf{92.2} & \cellcolor{tableblue}\textbf{80.3}
& \cellcolor{tableblue}\textbf{94.4} & \cellcolor{tableblue}\textbf{88.8} & \cellcolor{tableblue}\textbf{83.4}
& \cellcolor{lighttableblue}{96.4} & \cellcolor{lighttableblue}{92.2} & 86.9
& \cellcolor{tableblue}\textbf{98.2} & \cellcolor{tableblue}\textbf{96.5} & \cellcolor{tableblue}\textbf{94.3} & 10.0 \\
\bottomrule[1.2pt]
\end{tabular}
}
\caption{\textbf{Pose tracking accuracy and runtime across four benchmarks.}
We report ADD-S and ADD (AUC~\%, $0$–$0.1$\,m), and BOP Average Recall (AR) on HOI4D, YCBInEOAT, HO3D, and YCB-Video (average sequence length shown under each dataset name), alongside average inference speed (fps) across all four datasets. AR is omitted for HOI4D, because of no BOP-format ground
truth available. Methods are grouped by category:
\textbf{CAD}: requires CAD models; \textbf{TbR}: tracking-by-reconstruction; \textbf{RPE}: relative pose estimation; \textbf{PGO}: pose-graph-optimization-based tracking, the category our method belongs to alongside BundleTrack. Best result among non-CAD methods per column in \textbf{bold}, second-best \cellcolor{lighttableblue}{underlined} (tied values are marked identically). Among non-CAD methods, our method leads on YCBInEOAT, YCB-Video, and HOI4D, and is comparable to BundleSDF on HO3D, while running substantially faster than other tracking-by-reconstruction and pose-graph-based baselines. BundleTrack~(R.): BundleTrack re-run with RoMa~v2 as the feature matcher, using the official codebase. $\dagger$: numbers taken from the original paper runtime reported on a different GPU than ours 
(\cref{tab:run_diff_gpus})}
\label{table:main_results}
\end{table*}

\label{sec:compare_sota}

We compare our method to state-of-the-art pose tracking methods on four benchmarks: HOI4D, YCBInEOAT, HO3D, and YCB Video in ~\cref{table:main_results}. For a fair comparison, we re-run BundleTrack with RoMa v2 as its feature matcher (BundleTrack (R.)), matching our own setting; RPE baselines estimate each frame's pose relative to the first frame.

Compared to FoundationPose, using CAD models, our method leads on three of four benchmarks on ADD metric. HOI4D is the hardest of our four benchmarks, where our method outperforms FoundationPose by a large margin. We hypothesize that this difference arises from how the two methods propagate information over time. While FoundationPose's tracking mode refines each pose from a small set of hypotheses conditioned only on the previous frame, our loop-closure edges reconnect to earlier frames and propagate reliable information using uncertainty.

Among non-CAD methods, we reach the best or near-best accuracy on three of four benchmarks. Only on HO3D, BundleSDF and BundleTrack (R.) lead, by under one point. Against BundleTrack (R.), we retain global consistency while replacing its dense alignment objective with a lightweight, uncertainty-aware pose-consistency objective, enabling substantially faster updates.

\noindent\textbf{HOI4D.} ~\cref{tab:hoi4d_full} analyzes HOI4D by looking into the performance of all rigid object categories: toy car (\faCarSide), kettle (\icon{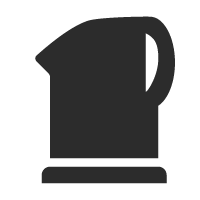}), knife (\icon{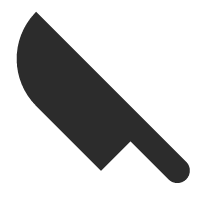}), chair (\faChair), and bowl (\icon{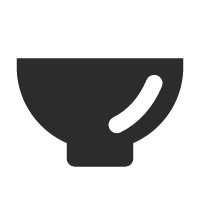}). 
UNOPose performs well since its training data is biased towards the large viewpoint changes present in the fast motion of HOI4D.
The ADD(-S) threshold (0.1\,m) is fixed across objects, which can compress rotation differences for small objects such as the toy car and knife.
We therefore provide the rotation only metric in \emph{supplementary material}.\begin{table}[t]
\centering
\footnotesize
\setlength{\tabcolsep}{4pt}
\resizebox{\linewidth}{!}{%
\begin{tabular}{c ccccc ccccc}
\toprule
\multirow{2}{*}{\textbf{Cat.}} & \multicolumn{5}{c}{\textbf{ADD (\%)}} & \multicolumn{5}{c}{\textbf{ADD-S (\%)}} \\
\cmidrule(lr){2-6} \cmidrule(lr){7-11}
& Ours & BS & BT & UP & FP & Ours & BS & BT & UP & FP \\
\midrule
\faCarSide  & \cellcolor{tableblue}\textbf{86.3} & 84.4 & {84.9} & 83.2 & 60.0 & \cellcolor{tableblue}\textbf{93.5} & 92.9 & 93.3 & 93.2 & 72.8 \\
\icon{icon/kettle-2.png} & \cellcolor{tableblue}83.3 & 83.3 & \textbf{83.5} & 80.9 & 78.3 & \cellcolor{tableblue}92.5 & 92.5 & \textbf{92.6} & 92.2 & 89.7 \\
\icon{icon/knife-fill.png}    & \cellcolor{tableblue}82.4 & 78.2 & 81.2 & \textbf{82.5} & 75.1 & \cellcolor{tableblue}91.7 & 87.5 & 90.9 & \textbf{92.2} & 84.8 \\
\faChair   & \cellcolor{tableblue}75.3 & 74.8 & 75.3 & 76.2 & \textbf{76.3} & \cellcolor{tableblue}88.9 & 87.6 & 88.1 & 89.2 & \textbf{89.3} \\
\icon{icon/Bowl.png}   & \cellcolor{tableblue}\textbf{74.3} & 74.1 & 74.2 & 73.9 & 42.6 & \cellcolor{tableblue}\textbf{94.4}& 93.5 & 93.9 & 94.1 & 85.4 \\
\midrule
Avg.      & \cellcolor{tableblue}\textbf{80.3} & 79.0 & 79.8 & 79.4 & 66.5 & \cellcolor{tableblue}\textbf{92.2} & 90.8 & 91.8 & \textbf{92.2} & 84.4 \\
\bottomrule
\end{tabular}%
}
\caption{\textbf{Per-category comparison on five HOI4D rigid object categories.} We report ADD-AUC and ADD-S-AUC (\%, $0$--$0.1$\,m). BS: BundleSDF, BT: BundleTrack, UP: UNOPose, FP: FoundationPose.}
\label{tab:hoi4d_full}
\end{table}
\begin{table}[ht]
\centering
\footnotesize
\setlength{\tabcolsep}{3.5pt}
\resizebox{\linewidth}{!}{%
\begin{tabular}{lccccc|c}
\toprule[1.2pt]
Matcher & Matching & Solve+Unc. & Frame Sel. & PGO & Pool Update & Ours (fps)\\
\midrule
RoMa v2 & 0.05 & 0.04 & 5e-4 & 2e-3 & 2e-4 & 10.8 \\
\bottomrule[1.2pt]
\end{tabular}
}
\caption{\textbf{Per-stage latency on YCB Video.} Results reported in seconds per frame and averaged over the dataset. The final column reports total tracking throughput in frames per second.} 
\label{tab:runtime_stages}
\end{table}

\begin{table}[ht]
    \centering
    \footnotesize
    \begin{tabular}{c c c}
    \toprule[1.2pt]
    \textbf{Method} & \textbf{GPU} & \textbf{fps} \\
    \midrule
    BundleTrack (LF-Net)~\cite{bundletrack}
        & RTX 2080 Ti & 10.0 \\
    6DOPE-GS~\cite{6dopegs}
        & RTX 4090 & 4.2 \\
    Ours & RTX 4090 & 9.0 \\
    \bottomrule[1.2pt]
    \end{tabular}%
    \captionof{table}{
    \textbf{Runtime on different GPUs.}
    Numbers are reported from the original papers.
    }
    \label{tab:run_diff_gpus}
\end{table}

\noindent\textbf{Runtime Analysis.} \cref{table:main_results} reports average runtime (fps) across all four benchmarks on the same H100 GPU (per-dataset breakdowns in \emph{supplementary material}). Our method runs at 10.0 fps on average, 5.5 to 7 times faster than the TbR (BundleSDF) and PGO baselines (BundleTrack (R.)) benchmarked on the same GPU. Native BundleTrack is reported at the same rate as our method, but uses a different GPU and matcher and is therefore not
directly comparable.  

\cref{tab:runtime_stages} breaks down per-frame runtime by stage on YCB Video. Matching dominates the cost ($\sim$0.05s per frame to be matched to $K=5$ selected keyframes), while our compact pose graph makes global consistency nearly free by comparison: frame selection, PGO, and memory pool update together add only 2.7\,ms, against 75\,ms for BundleTrack's dense-alignment PGO step under the same matcher and hardware (see \emph{supplementary material}). 
For real-world deployment without object masks, an additional per-frame segmentation stage can be pipelined into a separate CUDA stream, since it does not depend on our tracking backend. For reference, XMem~\cite{cheng2022xmem} runs at 100 fps, SAM2~\cite{sam2} runs at 50 fps for a $640\times480$ video stream. 

Finally, since 6DOPE-GS's code is unpublished, we report our runtime on its benchmarked GPU (RTX 4090) alongside their published numbers in \cref{tab:run_diff_gpus}. We also
include BundleTrack's reported runtime for reference, though it is likewise not
directly comparable, as we lack access to its benchmarked GPU.

\subsection{Sensitivity to Correspondence Quality}

\begin{table}[t]
\centering
\resizebox{\linewidth}{!}{%
\begin{tabular}{lc cc cc cc cc}
\toprule[1.5pt]
\multirow{2}{*}{\textbf{Method}}& \multirow{2}{*}{\textbf{Matcher}} & \multicolumn{2}{c}{\textbf{HOI4D}} & \multicolumn{2}{c}{\textbf{YCBInEOAT}} & \multicolumn{2}{c}{\textbf{HO3D}} & \multicolumn{2}{c}{\textbf{YCB Video}} \\
\cmidrule(lr){3-4} \cmidrule(lr){5-6} \cmidrule(lr){7-8} \cmidrule(lr){9-10}
& & ADD-S(\%) & ADD(\%) & ADD-S(\%) & ADD(\%) & ADD-S(\%) & ADD(\%) & ADD-S(\%) & ADD(\%)\\
\midrule
BundleTrack & \multirow{2}{*}{RoMa v2} & 91.8 & 80.2 & 93.8 & 88.0 & \textbf{96.5} & \textbf{92.6} & 97.5 & 94.9 \\
Ours        & & \cellcolor{tableblue}\textbf{92.2} & \cellcolor{tableblue}\textbf{80.3} & \cellcolor{tableblue}\textbf{94.4} & \cellcolor{tableblue}\textbf{88.8} & \cellcolor{tableblue}{96.4} & \cellcolor{tableblue}{92.2} & \cellcolor{tableblue}\textbf{98.2} & \cellcolor{tableblue}\textbf{96.5} \\
\midrule
BundleTrack & \multirow{2}{*}{LoFTR}   & 92.0   & 80.1   & 92.5 & 84.9 & 94.1 & 79.3 & 97.3 & 93.9 \\
Ours &    & \cellcolor{tableblue}\textbf{92.2}   & \cellcolor{tableblue}\textbf{80.6}   & \cellcolor{tableblue}\textbf{93.0} & \cellcolor{tableblue}\textbf{87.8} & \cellcolor{tableblue}\textbf{94.8} & \cellcolor{tableblue}\textbf{85.7} & \cellcolor{tableblue}\textbf{97.4} & \cellcolor{tableblue}\textbf{94.2}\\
\bottomrule[1.5pt]
\end{tabular}%
}
\caption{\textbf{Comparison to dense alignment-PGO method BundleTrack.} We show the direct comparison between dense-alignment and pose consistency optimization under the same correspondence estimator. The pose-level optimization preserves accuracy while dramatically reducing optimization cost.
}
\label{tab:fm_results}
\end{table}
\cref{tab:fm_results} directly compares our method to the dense alignment PGO method (BundleTrack) with the same feature matcher. With the state-of-the-art dense 2D feature matcher, RoMa v2, the two objectives perform comparably. To test this under lower-quality correspondences, we repeat the comparison with LoFTR, whose smaller network architecture and training data scale make it a weaker matcher. With LoFTR, our method leads on all datasets and both metrics, most notably by 6.4 ADD points on HO3D. This suggests our compact pose graph formulation degrades more gracefully than the prior method with dense-alignment objective as correspondence quality decreases. A fuller comparison, including frame-to-frame baselines and BundleTrack using native LF-Net~\cite{ono2018lf} matcher, is in the \emph{supplementary material}. 

\begin{table}[t]
\centering
\footnotesize
\setlength{\tabcolsep}{6pt}
\resizebox{\linewidth}{!}{%
\begin{tabular}{l cc cc cc cc}
\toprule[1.2pt]
\multirow{2}{*}{\textbf{Uncertainty}} & \multicolumn{2}{c}{\textbf{HOI4D}} & \multicolumn{2}{c}{\textbf{YCBInEOAT}} & \multicolumn{2}{c}{\textbf{HO3D}} & \multicolumn{2}{c}{\textbf{YCB Video}} \\
\cmidrule(lr){2-3} \cmidrule(lr){4-5} \cmidrule(lr){6-7} \cmidrule(lr){8-9}
& ADD-S(\%) & ADD(\%)  & ADD-S(\%) & ADD(\%)  & ADD-S(\%) & ADD(\%)  & ADD-S(\%) & ADD(\%)  \\
\toprule
\cellcolor{tableblue}Identity (\emph{RoMa v2})
& 92.11 & 80.33 & 93.69 & 87.60 & 96.14 & 90.77 & 96.82 & 94.19 \\

\cellcolor{lighttableblue}Trace
& \gain{-0.28} & \gain{-0.23} & \gain{-1.15} & \gain{-3.07}
& \textbf{\gain{+0.39}} & \textbf{\gain{+1.80}}
& \textbf{\gain{+1.34}} & \gain{+2.24} \\

\cellcolor{lighttableblue}Covariance
& \textbf{\gain{+0.09}} & \textbf{\gain{+0.00}}
& \textbf{\gain{+0.70}} & \textbf{\gain{+1.20}}
& \gain{+0.26} & \gain{+1.43}
& \textbf{\gain{+1.34}} & \textbf{\gain{+2.31}} \\

\cellcolor{lighttableblue}\color{gray} Oracle\*
& \color{gray}+0.40 & \color{gray}+1.56 & \color{gray}+1.08 & \color{gray}+2.40
& \color{gray}+0.80 & \color{gray}+2.95 & \color{gray}+0.17 & \color{gray}+1.20 \\
\midrule

\cellcolor{tablelightyellow}Identity (\emph{LoFTR})
& 91.40 & 79.35 & 92.38 & 83.52 & 94.07 & 83.34 & 96.03 & 92.30 \\

\cellcolor{tableyellow}Trace
& +0.33 & +0.68 & \textbf{+1.21} & {+3.40}
& \textbf{+1.42} & \textbf{+3.24}
& \textbf{+1.92} & \textbf{+2.92} \\

\cellcolor{tableyellow}Covariance
& \textbf{+0.80} & \textbf{+1.25} & +0.62 & +\textbf{4.28}
& +0.78 & +2.36 & +1.37 & +1.90 \\

\cellcolor{tableyellow}\color{gray} Oracle\*
& \color{gray}+1.12 & \color{gray}+2.69 & \color{gray}+1.77 & \color{gray}+5.00 & \color{gray}+2.34 & \color{gray}+7.76 & \color{gray}+0.11 & \color{gray}+2.69 \\
\bottomrule[1.2pt]
\end{tabular}
}
\caption{\textbf{Ablation on uncertainty structures.} We compare four
levels of uncertainty information used in pose graph optimization:
\textit{Identity} (no per-edge uncertainty), \textit{Trace} (isotropic,
using only the scalar magnitude of our estimated covariance), \textit{Covariance} (our full anisotropic per-edge estimate), and \textit{Oracle*} (derived from ground-truth pose error, an upper bound). }
\label{tab:uncertainty_ablation}
\end{table}


\subsection{Drift Analysis}
\begin{figure}[ht]
  \centering
   \includegraphics[width=\linewidth]{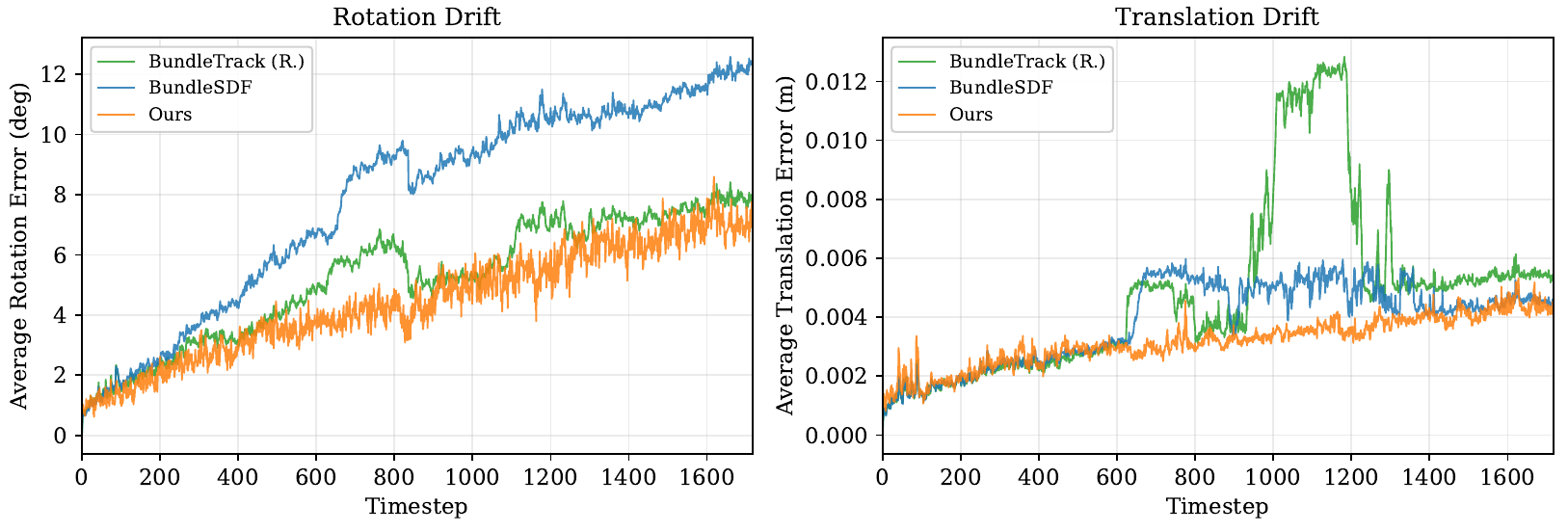}
    \caption{\textbf{Drift analysis on YCB Video dataset.} Rotation and
    translation drift over the full 1718-timestep sequence.}
  \label{fig:drift}
\end{figure}

We evaluate drift on YCB Video, the longest sequence among our four benchmarks (Table~\ref{table:main_results}), making it the most direct real-data test of long-horizon stability in our evaluation suite. Figure~\ref{fig:drift} shows rotation and translation drift over the full sequence. Our method maintains the lowest mean error on both axes throughout, and, critically, shows no sustained failure episode over the full run.

\subsection{Ablation Studies}

We ablate the estimated uncertainty along three axes: how much of its structure is used, how it behaves as correspondence quality degrades, and how well it predicts the true error.

\begin{figure}
    \centering
   \begin{tabular}{@{}p{0.3\linewidth}<{\centering}p{0.3\linewidth}<{\centering}p{0.3\linewidth}<{\centering}@{}}
    \footnotesize Keyframe & \footnotesize Current Frame & \footnotesize Estimated Pose
    \end{tabular}
    \includegraphics[width=\linewidth]
    {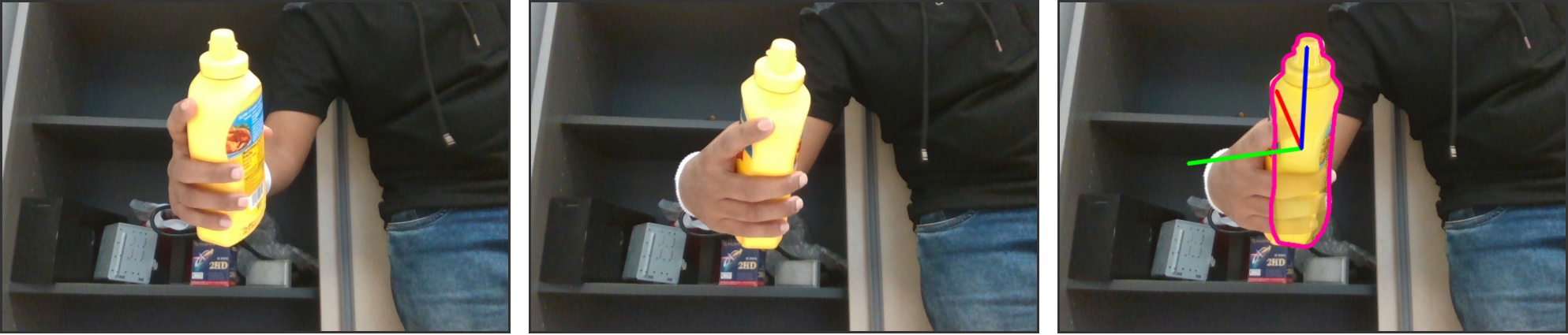}
    \begin{tabular}{ll}
   
     \footnotesize Rot. Error: 37.66\degree &   \footnotesize Uncertainty: 1.22 \\    
     \footnotesize Tran. Error: 0.85\,cm      &  \footnotesize Uncertainty: 0.81   
    \end{tabular}

   \includegraphics[width=\linewidth]{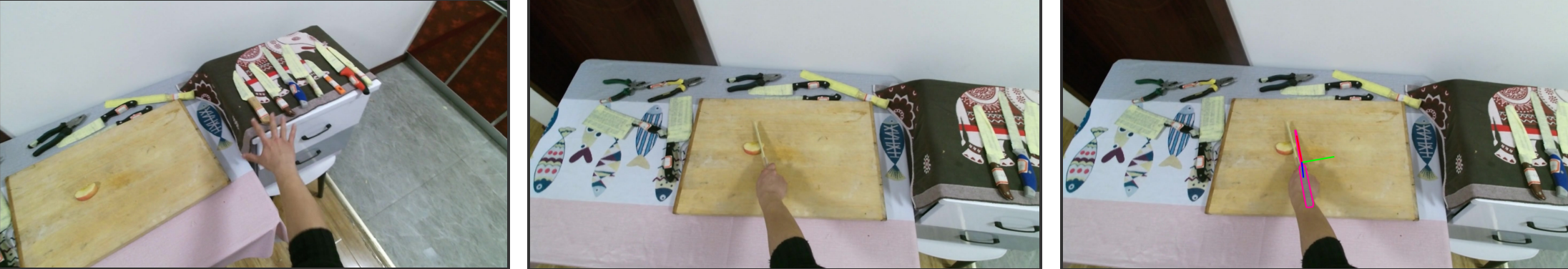}
    \footnotesize
    \begin{tabular}{ll}
    \footnotesize Rot. Error: 176.96\degree & Uncertainty: 36.92 \\
    Tran. Error: 7.69\,cm      & Uncertainty: \hphantom{0}1.64    
    \end{tabular}

    \caption{\textbf{Overconfident uncertainty for heavy occluded symmetric objects and thin objects from end-on views.} Top: under the heavy occlusion, the hand covers the label, leaving only correspondences on the rotationally-symmetric body; Bottom: thin object viewed from its end-on view axis, leaving little visual signal. For the purpose of visualization, we show the square root of the covariance's trace as a scalar uncertainty. }
    \label{fig:uncertainty_failure}
\end{figure}

\noindent\textbf{Uncertainty Structure.} 
~\cref{tab:uncertainty_ablation} compares four levels of uncertainty information used in pose-graph optimization: \textit{Identity} (no per-edge uncertainty), \textit{Trace} (isotropic, using only the scalar magnitude of our estimated covariance), \textit{Covariance} (our full anisotropic per-edge estimate, \cref{eq:gn_info,eq:jac}), and \textit{Oracle} (derived from pose error between estimated and ground truth, as an upper bound). 
Both the trace and the full
covariance improve accuracy over unweighted edges.
On two of the four benchmarks,
however, the directional information in the full covariance brings no consistent
benefit over purely magnitude-based weighting, which suggests that an anisotropic
covariance only helps when its estimated directions are correct.
The example in \cref{fig:uncertainty_failure} (Top) shows an edge where this fails: the estimation is well-calibrated in translation but severely overconfident along rotation.

\noindent\textbf{Uncertainty vs. Correspondence Quality.} The lower block of \cref{tab:uncertainty_ablation} repeats the comparison with LoFTR in place of RoMa v2. The gain from covariance weighting over unweighted edges is larger with the weaker matcher, indicating that the estimated uncertainty matters most when the underlying relative poses are least reliable. This mirrors the trend in \cref{tab:fm_results}: with high-quality correspondences our compact formulation already matches the dense-alignment objective, and the uncertainty is what limits the influence of bad edges as correspondence quality drops.

\begin{figure}
    \centering
    \includegraphics[width=1.0\linewidth]{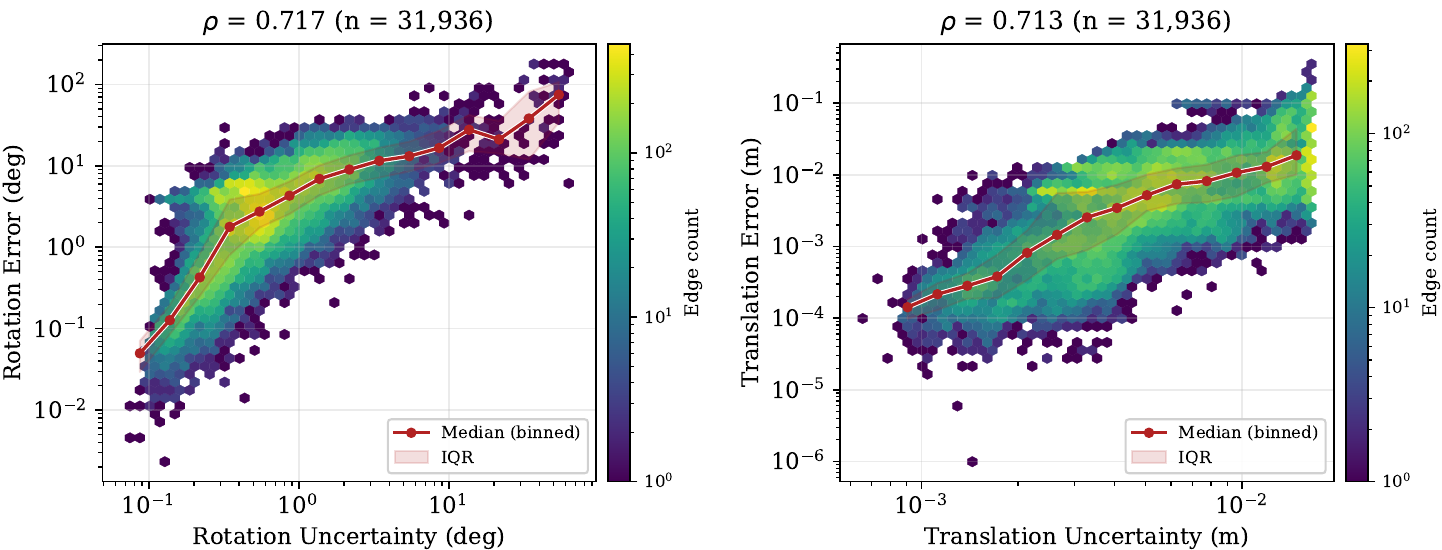}
    \includegraphics[width=1.0\linewidth]{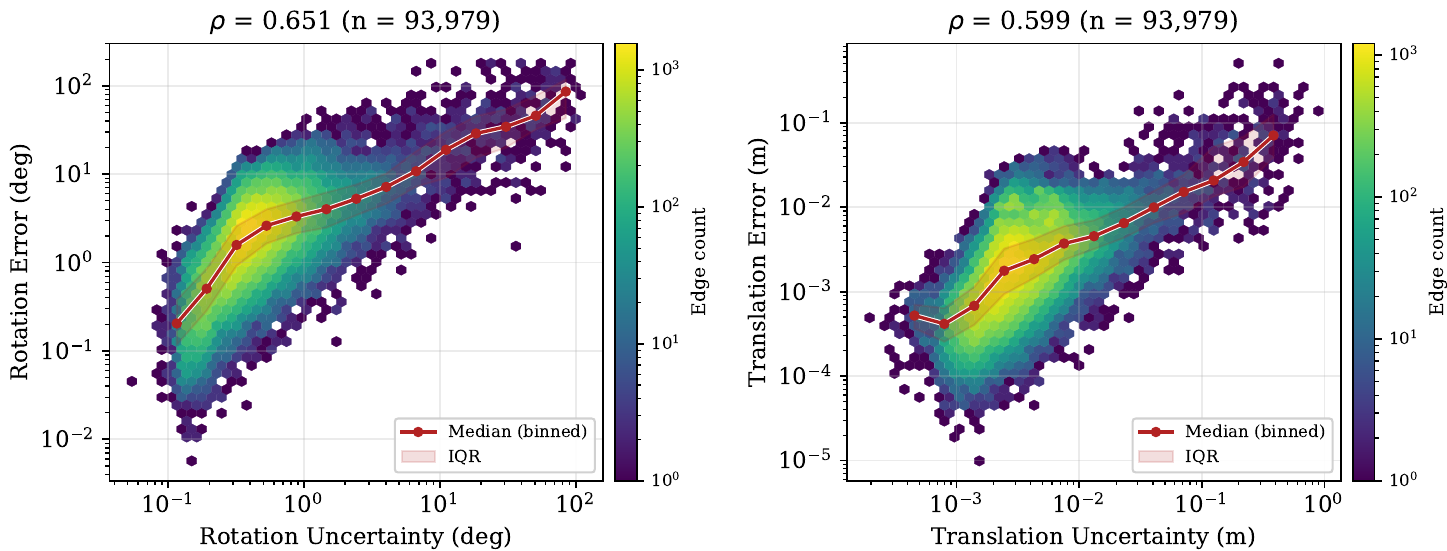}
    \caption{\textbf{Uncertainty vs. Pose Error.} We show the joint distribution of estimated uncertainty and ground-truth pose error, for rotation (left) and translation (right), on HO3D and YCBInEOAT. Hexbin color indicates edge density on a log scale; the red line and shaded band show the median and interquartile range of error within log-spaced uncertainty bins. 
    }
    \label{fig:uncertainty_calibration}
\end{figure}

\noindent\textbf{Uncertainty--Error Correlation.} We evaluate whether our estimated uncertainty predicts the true pose error by computing the Spearman rank correlation between the trace of the estimated per-edge uncertainty and the ground-truth error on HO3D and YCBInEOAT (\cref{fig:uncertainty_calibration}). The positive correlation indicates that the predicted uncertainty is broadly associated with the true error.

\noindent\textbf{Failure Cases.} We observe two main failure cases: symmetric objects with weak textures under heavy occlusions (\cref{fig:uncertainty_failure} Top), and thin objects viewed from end-on views (\cref{fig:uncertainty_failure} Bottom). We refer readers to the \emph{supplementary material} for more quantitative results on uncertainty-weighted PGO and performance under challenging cases.

\noindent\textbf{Keyframe Selection.}
~\cref{tab:keyframe_sel} ablates the two components of our keyframe selection stage: the memory pool and farthest-frame sampling (FFS). 
Removing the memory pool allows loop closure edges to be drawn from all past frames rather than from a small set of viewpoint-diverse ones, so the selected frames become redundant and carry less information.
Replacing farthest frame sampling with the adjacent $K$ frames leads to severe pose degeneration since graph connectivity becomes purely local. This confirms that the value of loop closure comes from spanning the sequence rather than from the number of edges.
\begin{table}[ht]
    \centering
    \footnotesize
    \begin{tabular}{lcccc}
    \toprule[1.2pt]
    \multirow{2}{*}{\textbf{Keyframe}} & \multicolumn{2}{c}{\textbf{HO3D}}  & \multicolumn{2}{c}{\textbf{YCBInEOAT}}\\
     \cmidrule(lr){2-3}  \cmidrule(lr){4-5} 
     & {\scriptsize ADD-S (\%)} & {\scriptsize ADD (\%)} & {\scriptsize ADD-S (\%)} & {\scriptsize ADD (\%)} \\
    \midrule
    Ours     & {96.37} & {92.23} & {94.18} & {88.75} \\
    w.o FFS  & \gain{-4.22} & \gain{-14.10} & \gain{-1.71} & \gain{-3.81} \\
    w.o Pool & \gain{-0.10} & \gain{-0.29}  & \gain{-0.33} & \gain{-0.56} \\
    \bottomrule[1.2pt]
     \end{tabular}%
    \caption{\textbf{Ablation on keyframe selection.} Ours w.o. FFS use the direct K neighbors to build loop closure edges; Ours w.o. memory pool allows loop-closure candidates to be drawn from all past frames.}
    \label{tab:keyframe_sel}
\end{table} 
\section{Conclusion}
In this work, we propose a lightweight, correspondence-agnostic framework for online 6D pose tracking of novel objects, built on a compact pose graph optimized with iSAM2. 
Rather than fusing dense correspondences into the objective, we use them only to produce relative pose measurements, which the pose graph combines with geometry-derived per-edge uncertainty and viewpoint-diverse frame selection.
Across four benchmarks, our method reaches near-best accuracy among non-CAD methods at a fraction of the optimization cost of reconstruction- and pose-graph-based baselines, while remaining stable over sequences of more than a thousand frames. Our results suggest that accurate object tracking does not require retaining dense correspondences throughout optimization. Instead, compact relative pose measurements, weighted by how well each is constrained, provide comparable accuracy at substantially lower cost.
\section{Limitation}
Our method has two main limitations. First, we assume the object is rigid and we leave tracking articulated and deformable objects as future work. Second, our covariance is a local approximation around the solution RANSAC-Umeyama converges to, so it can capture a flat, poorly-constrained direction but not a well-separated alternative solution (e.g. for textureless symmetric objects). Further work could be a learning-based uncertainty estimator for 6D poses to represent multi-modal pose distributions.

{
    \small
\bibliographystyle{ieeenat_fullname}
    \bibliography{main}
}

\clearpage
\setcounter{page}{1}
\maketitlesupplementary

\section{Overview}
In this supplementary document, we provide additional details and analyses that complement the main paper. We first describe implementation details, including data preprocessing and the hyperparameters used in our method (Sec.~\ref{sec:implementation}). We then derive the $SE(3)$ pose uncertainty (Sec.~\ref{sec:derivation}). Next, we review the evaluation metrics for pose tracking quality (Sec.~\ref{sec:metrics}). We show the comparison to different tracking approaches (\cref{sec:fm}) and runtime breakdown (\cref{sec:runtime}). Furthermore, we include detailed statistics of HOI4D dataset (\cref{sec:sota_detail}). Finally, we provide additional qualitative results on uncertainty-aware PGO (\cref{sec:uncertainty_pgo}) and challenge cases (\cref{sec:challenges}).

\section{Implementation Details}
\label{sec:implementation}

\subsection{Input Preprocessing}
We follow standard input preprocessing used in prior RGB-D tracking pipelines~\cite{bundlesdf,bundletrack}. 
For object masks, we use XMem~\cite{cheng2022xmem} masks on HO3D~\cite{ho3d} and ground-truth masks on other datasets. 
For correspondence estimation, we crop object-centric image regions and resize them to a fixed resolution: $320 \times 320$ for RoMa v2 and $400 \times 400$ for LoFTR.

\subsection{Hyperparameters}
For pose graph optimization, we use the Cauchy robust loss in GTSAM~\cite{gtsam} with scale parameter $k{=}1.0$. 
We set the frame selection threshold to $\tau_k{=}25^\circ$ for RoMa v2 and $\tau_k{=}15^\circ$ for LoFTR. We use 1000 correspondences for both matchers in our method. For memory pool update, we set the threshold $\tau_m = 5\degree$. We set the RANSAC inlier threshold to $d_{r} = 2e-3$.  These parameters are applied to all benchmarks where we evaluate our method, without tuning.

\subsection{Ablation of Parameters}
We ablate $\tau_k$ for RoMa v2 and LoFTR. We observe that LoFTR is more sensitive to larger $\tau_k$, while RoMa v2 is more robust to larger $\tau_k$. The difference could come from the generalization of different methods to large viewpoint changes. RoMa v2 shows stronger performance in terms of handling such cases. However, for both cases, too small a $\tau_k$ also suppresses the performance due to similar views that do not provide much additional information. 
\begin{figure}
    \centering
    \includegraphics[width=0.8\linewidth]{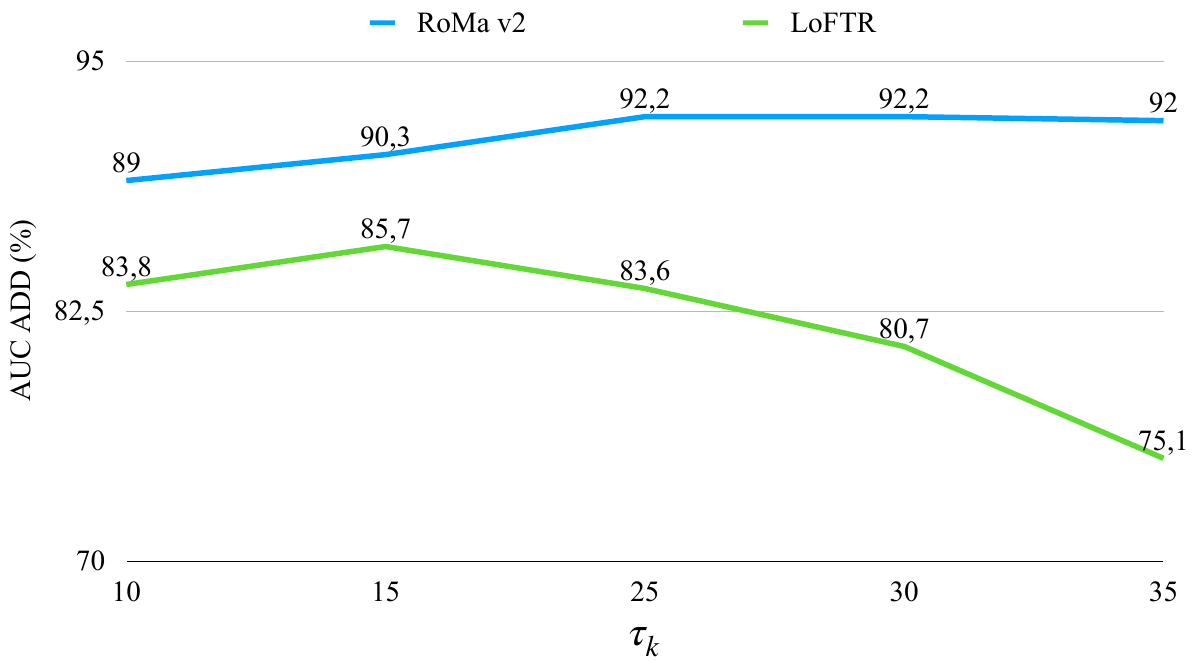}
    \caption{\textbf{Parameter Sweep of $\tau_k$ for different feature matchers.}}
    \label{fig:placeholder}
\end{figure}

\section{Uncertainty Estimation}
\label{sec:derivation}
In this section, we show how we derive the per-edge covariance $\Sigma_{ik}$. We use RANSAC Umeyama~\cite{umeyama1991least} to identify the inlier set of the correspondences $\mathcal{C}_{ik}^{in}$.  For simplicity, we omit the frame indices $ik$ here. Our goal is to estimate the covariance $\Sigma \in \mathbb{R}^{6\times6}$ that describes the uncertainty of geometric alignment from the target function,
\begin{equation}
\min_{R\in SO(3),\,t\in\mathbb{R}^3}
\sum_{i=1}^{N} w_i \,\left\| Rb_i + t - a_i \right\|_2^2, a_i \in \mathbb{R}^3,b_i\in \mathbb{R}^{3},
\end{equation}
where $N$ is the cardinality of the inlier set $|\mathcal{C}^{\text{in}}|.$ For each correspondence, we define the weighted residual
\begin{equation}
e_i(R,t) \triangleq \sqrt{w_i}\,(Rb_i+t-a_i)\in\mathbb{R}^3.
\end{equation}
We perturb the rotation on $SO(3)$ as
\begin{equation}
R(\delta\theta)=R\exp([\delta\theta]_\times),
\end{equation}
where $\delta\theta\in\mathbb{R}^3$ is a small rotation vector. Let
\begin{equation}
r_i(\delta)=R\exp([\delta\theta]_\times)b_i+t-a_i,
\end{equation}
with $u\in\mathbb{R}^3$ an arbitrary direction. Using Rodrigues' formula,
\begin{equation}
\exp([\delta\theta]_\times)=I+\sin(\delta)[\theta]_\times+(1-\cos \delta)[\theta]_\times^2,
\end{equation}
we obtain
\begin{equation}
r_i'(\delta)=R\left(\cos(\delta)[\theta]_\times+\sin(\delta)[\theta]_\times^2\right)b_i.
\end{equation}
Evaluating at $\delta=0$ gives
\begin{equation}
r_i'(0)=R[\theta]_\times b_i = -R[b_i]_\times \theta.
\end{equation}
Therefore, the directional derivative of the weighted residual is
\begin{equation}
e_i'(0)=\sqrt{w_i}\,r_i'(0) = -\sqrt{w_i}\,R[b_i]_\times \theta.
\end{equation}
Since $e_i'(0)=J_{i,\mathrm{rot}}\,\theta$ for arbitrary $\theta$, the Jacobian with respect to the rotation perturbation is
\begin{equation}
J_{i,\mathrm{rot}}=-\sqrt{w_i}\,R[b_i]_\times.
\end{equation}
Similarly, the Jacobian with respect to translation is
\begin{equation}
J_{i,\mathrm{trans}}=\sqrt{w_i}\,I_3.
\end{equation}
Thus, the full Jacobian for the $i$-th correspondence is
\begin{equation}
J_i= \sqrt{w_i}\,\begin{bmatrix}
-R[b_i]_\times & I_3
\end{bmatrix}\in\mathbb{R}^{3\times 6}.
\end{equation}
Stacking all correspondences yields the Gauss--Newton approximation
\begin{equation}
H \approx \sum_{i=1}^N J_i^\top J_i.
\end{equation}
To calibrate the correct scale of $\Sigma$, we apply the residual-variance factor to align the uncertainty with the real scale as follows:
\begin{equation}
\label{eq:sigma_cal}
\hat\sigma^2 = 
\frac{\sum_{n=1}^{N} w_i\|r_i\|_2^2}
{\max(3|\mathcal{C}^{\text{in}}| - 6,\ 1)}.
\end{equation}
Last, reinstating frame indices yields $\Sigma_{ik} = \sigma^2_{ik}H^{-1}$

\section{Metrics}
\label{sec:metrics}
We follow the evaluation protocol used in previous work~\cite{bundlesdf,bundletrack,6dopegs} to assess 6-DoF object pose tracking quality.  For each video, we align the predicted trajectory to the ground-truth coordinate frame using the relative transformation between the predicted and ground-truth pose at the first frame. 
We report the area under the curve (AUC) of the \textit{ADD} and \textit{ADD-S} metrics, computed over distance thresholds in the range $[0,0.1]$\,m.

Given the ground-truth pose $(R,t)$, the estimated pose $(\tilde{R},\tilde{t})$, and the object model point set $\mathcal{M}$, the metrics are defined as:
\begin{equation}
    \text{ADD}=\frac{1}{|\mathcal{M}|}\sum_{x\in \mathcal{M}}
    \left\| (Rx+t)-(\tilde{R}x+\tilde{t}) \right\|_2,
\end{equation}
\begin{equation}
    \text{ADD-S}=\frac{1}{|\mathcal{M}|}\sum_{x_1\in \mathcal{M}}
    \min_{x_2\in \mathcal{M}}
    \left\| (Rx_1+t)-(\tilde{R}x_2+\tilde{t}) \right\|_2.
\end{equation}
Compared to ADD, ADD-S replaces one-to-one point correspondences with a closest-point distance and is therefore more suitable for symmetric objects, since there is symmetry ambiguity.

\section{Comparison to Different Tracking Approaches}
\label{sec:fm}
\cref{table:fm_detailed_results} compares our full method against direct frame-to-frame (F2F) tracking and BundleTrack~\cite{bundletrack}. For F2F setting, the pose of each query image is estimated relative to the previous frame alone, using three correspondence estimators~\cite{romav2,lofter,unopose}. Among the F2F results, RoMa v2 and LoFTR are considerably more robust and stable than UNOPose (\cref{table:fm_detailed_results} top). Pairing our tracking module with each backend (\cref{table:fm_detailed_results} bottom) improves accuracy consistently, and the size of the gain scales with how weak the backend is on its own.

\begin{table}[t]
\centering
\setlength\tabcolsep{3pt}
\resizebox{\linewidth}{!}{%
\begin{tabular}{l c c cc cc}
\toprule[1.2pt]
\multirow{2}{*}{Setting} & \multirow{2}{*}{PGO Target} & \multirow{2}{*}{RPE} & \multicolumn{2}{c}{HO3D} & \multicolumn{2}{c}{YCBInEOAT}\\
\cmidrule(lr){4-5} \cmidrule(lr){6-7} 
& & &{\scriptsize ADD-S (\%)} & {\scriptsize ADD (\%)} & {\scriptsize ADD-S (\%)} & {\scriptsize ADD (\%)}\\
\midrule
\multirow{3}{*}{\shortstack[l]{F2F \\ Tracking}} & 
\multirow{3}{*}{\shortstack[l]{-}} 
& UNOPose  & 20.1 & 5.8  & 22.1 & 10.2 \\
& & LoFTR    &   63.4   &   30.5   &   36.7   &   24.1  \\
& & RoMa v2   & 78.0 & 56.7 &   70.8   &   52.5  \\

\midrule
\multirow{3}{*}{\shortstack[l]{BundleTrack}} & 
\multirow{3}{*}{\shortstack[l]{\emph{Dense-Alignment}}}
& LFNet  &  92.4  & 66.0 & 93.0 & 87.3 \\ 
& & LoFTR   & 94.1  & 79.3 & 92.5 & 84.9  \\
& & RoMa v2  & \textbf{96.5} & \textbf{92.6} & 93.8 & \underline{88.0} \\ 

\midrule
\multirow{3}{*}{\shortstack[l]{Ours}} &
\multirow{3}{*}{\shortstack[l]{\emph{Pose-Consistency}}}
& UNOPose  &   86.4   &   61.6   &   \textbf{95.3}
 & 87.8 \\ 
& & LoFTR    &  {94.8} & {85.7}
    &   93.0   & 87.8 \\ 
& & RoMa v2  & \underline{96.4} & \underline{92.2} & {94.4} & \textbf{88.8} \\ 

\bottomrule[1.2pt]
\end{tabular}%
}
\caption{\textbf{Comparison of tracking approaches on HO3D and YCBInEOAT with different correspondence and pose estimation methods.} 
\textit{Top}: direct frame-to-frame (F2F) tracking using each correspondence/relative pose method alone. 
\textit{Middle}: BundleTrack with different feature matchers.
\textit{Bottom}: our online pose graph optimization (PGO) module paired with different pose estimators. As visible, our module provides a significant accuracy gain in all cases.
}
\label{table:fm_detailed_results}
\vspace{-0.5cm}
\end{table}

\section{Runtime Analysis}
\label{sec:runtime}
\noindent\textbf{Detailed Runtime.} \cref{tab:efficiency_per_dataset} provides a detailed runtime of our method and baselines on four benchmarks. 

\begin{figure}
    \centering
    \includegraphics[width=1.0\linewidth]{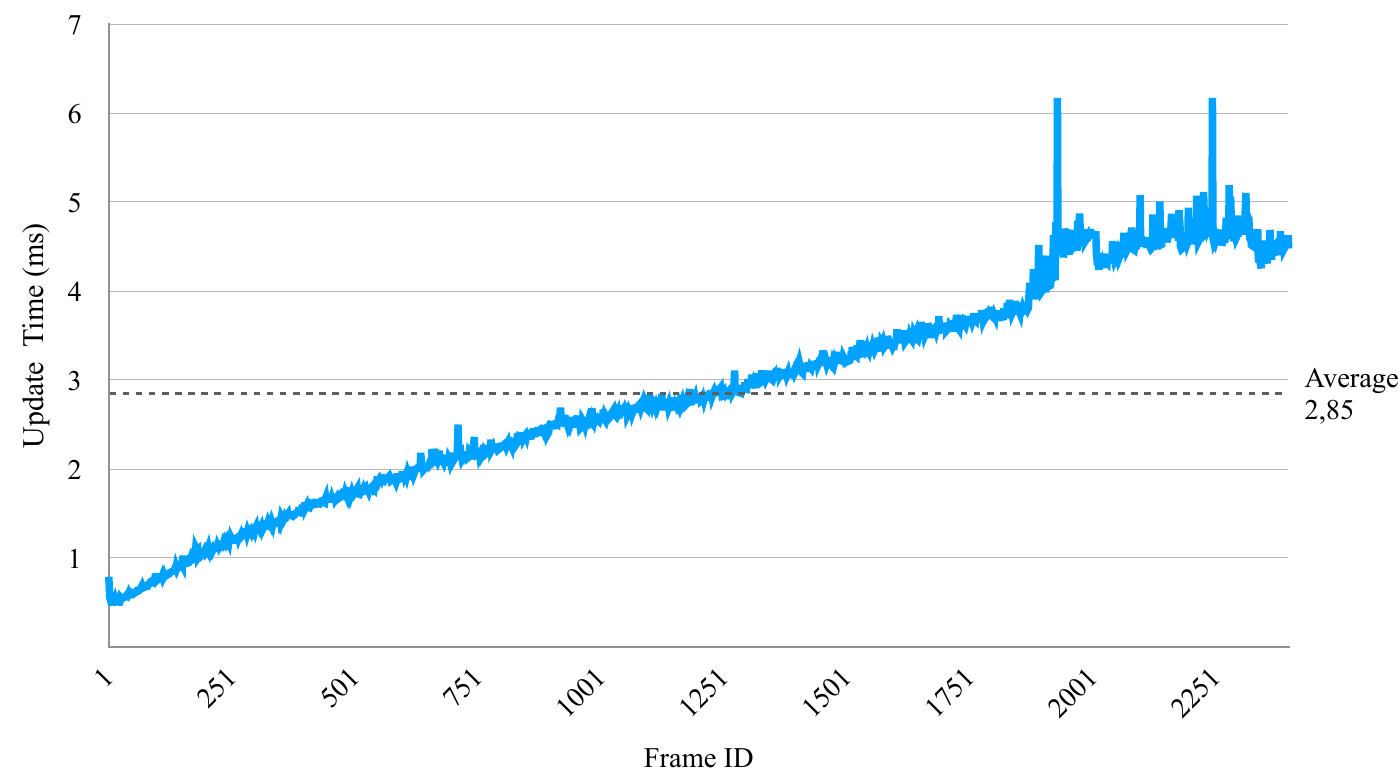}
     \captionof{figure}{
     \textbf{Relinearization time of iSAM2.} Result reports the time in milliseconds per frame averaged over frames on the YCB-Video dataset.}
    \label{fig:isam2}
\end{figure}

\noindent\textbf{iSAM2 Relinearization Time.} \cref{fig:isam2} shows iSAM2's relinearization time over the YCB Video sequence (1718 frames, our longest): cost grows gradually but stays low-millisecond throughout.

\begin{figure}
    \centering
    \includegraphics[width=\linewidth]{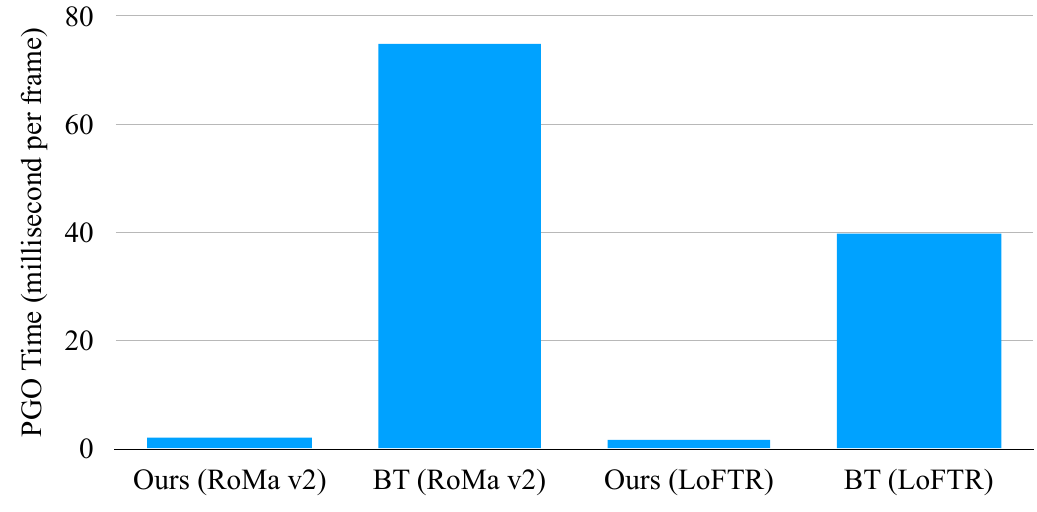}
    \caption{\textbf{PGO runtime compared to BundleTrack on YCB Video dataset.} We use the same number of correspondences for both method. Runtime is reported in millisecond per frame. }
    \label{fig:pgo_compare}
\end{figure}

\noindent\textbf{BundleTrack PGO Runtime.} 
\cref{fig:pgo_compare} reports our compact PGO update time (millisecond per frame) compared to dense alignment PGO method (BundleTrack~\cite{bundletrack}) on YCB Video dataset. For a fair comparison, we report both methods using the same feature matcher and the same number of correspondences.  Our compact PGO updates at a significantly faster speed, around 2 milliseconds per frame on average. 

\begin{table}[t]
\centering
\setlength{\tabcolsep}{6pt}
\resizebox{\linewidth}{!}{%
\begin{tabular}{l c c c c c c}
\toprule[1.2pt]
Method & GPU & HOI4D & YCBInEOAT & HO3D & YCB-Video & Average\\
\midrule
One2Any & \multirow{5}{*}{H100} & 30.0 & 30.0 & 30.0 & 30.0 & 30.0\\
FoundationPose (Tracking) & 
    & 10.0 & 33.0 & 18.3 & 30.1 & 22.9\\
UNOPose
    & & 10.0 & 10.0 & 10.0 & 10.0 & 10.0\\

BundleSDF
    & & 1.4 & 1.3 & 0.9 & 3.5 & 1.8\\
BundleTrack(R)
    & & 1.2 & 1.3 & 0.6 & 2.6 & 1.4\\
\midrule
Ours (RoMa v2)
    & H100 &  9.8 & 10.0 & 9.4 & 10.8 & 10.0\\
\bottomrule[1.2pt]
\end{tabular}
}
\caption{\textbf{Efficiency comparison across benchmarks.} Frame rate (fps)
per method, reported per dataset rather than pooled, since sequence length
and scene composition vary across benchmarks and can shift per-frame cost.}
\label{tab:efficiency_per_dataset}
\end{table}

\section{Results on HOI4D}
\label{sec:sota_detail}
\cref{tab:hoi4d_10} provides per-category results on rotation metric AUC@10$\degree$ on the HOI4D dataset compared to the state-of-the-art methods.
\begin{table}[t]
\centering
\footnotesize
\setlength{\tabcolsep}{6pt}
\resizebox{\linewidth}{!}{%
\begin{tabular}{l cccccc}
\toprule[1.2pt]
\multirow{2}{*}{\textbf{Method}} &  \multicolumn{5}{c}{\textbf{AUC@10\degree}} & \multirow{2}{*}{\textbf{Avg.}} \\
\cmidrule(lr){2-6}
& \faCarSide & \icon{icon/kettle-2.png} & \icon{icon/knife-fill.png} & \faChair & \icon{icon/Bowl.png} & \\
\midrule
FoundationPose(CAD)      & 25.5             & 3.6             & 41.1             & 20.6            & 70.5 & 32.2 \\
\midrule
BundleSDF      & 41.3          & 47.2 & 39.1         & 68.4          & 21.4 & 43.5 \\
UNOPose & 35.9       & 42.9         & 37.1         & 68.7         & 19.1 & 40.8 \\
BundleTrack (R.)      & 41.0          & \textbf{47.4}          & 40.5          & \textbf{70.4}         & 21.5 & 44.2 \\
Ours    & \cellcolor{tableblue}\textbf{46.6} & \cellcolor{tableblue}46.5     &\cellcolor{tableblue}\textbf{47.0} &\cellcolor{tableblue}70.0 & \cellcolor{tableblue}\textbf{22.2} & \cellcolor{tableblue}\textbf{46.5} \\
\bottomrule[1.2pt]
\end{tabular}
}
\caption{\textbf{Per-category rotation accuracy on five HOI4D rigid object categories.} We report rotation error AUC@$10^\circ$ (higher is better). Our method leads on three of five non-CAD categories and on average, most notably on knife and toy car.}
\label{tab:hoi4d_10}
\end{table}

\section{Qualitative Results of Uncertainty-Aware PGO}
\label{sec:uncertainty_pgo}
\begin{figure*}[t]
    \centering
   \begin{tabular}{@{}p{0.3\linewidth}<{\centering}p{0.3\linewidth}<{\centering}p{0.3\linewidth}<{\centering}@{}}
    \footnotesize \textbf{3D Trajectory} & \footnotesize \textbf{GT Pose} & \footnotesize \textbf{Estimated Pose}
    \end{tabular}
    
    \begin{subfigure}{\linewidth}
        \centering
        \begin{tabular}{
            @{}>{\centering\arraybackslash}m{0.12\linewidth}
               >{\centering\arraybackslash}m{0.20\linewidth}
               >{\centering\arraybackslash}m{0.62\linewidth}@{}
        }
            \textbf{w/o uncertainty} &
            \includegraphics[width=\linewidth, valign=c]
            {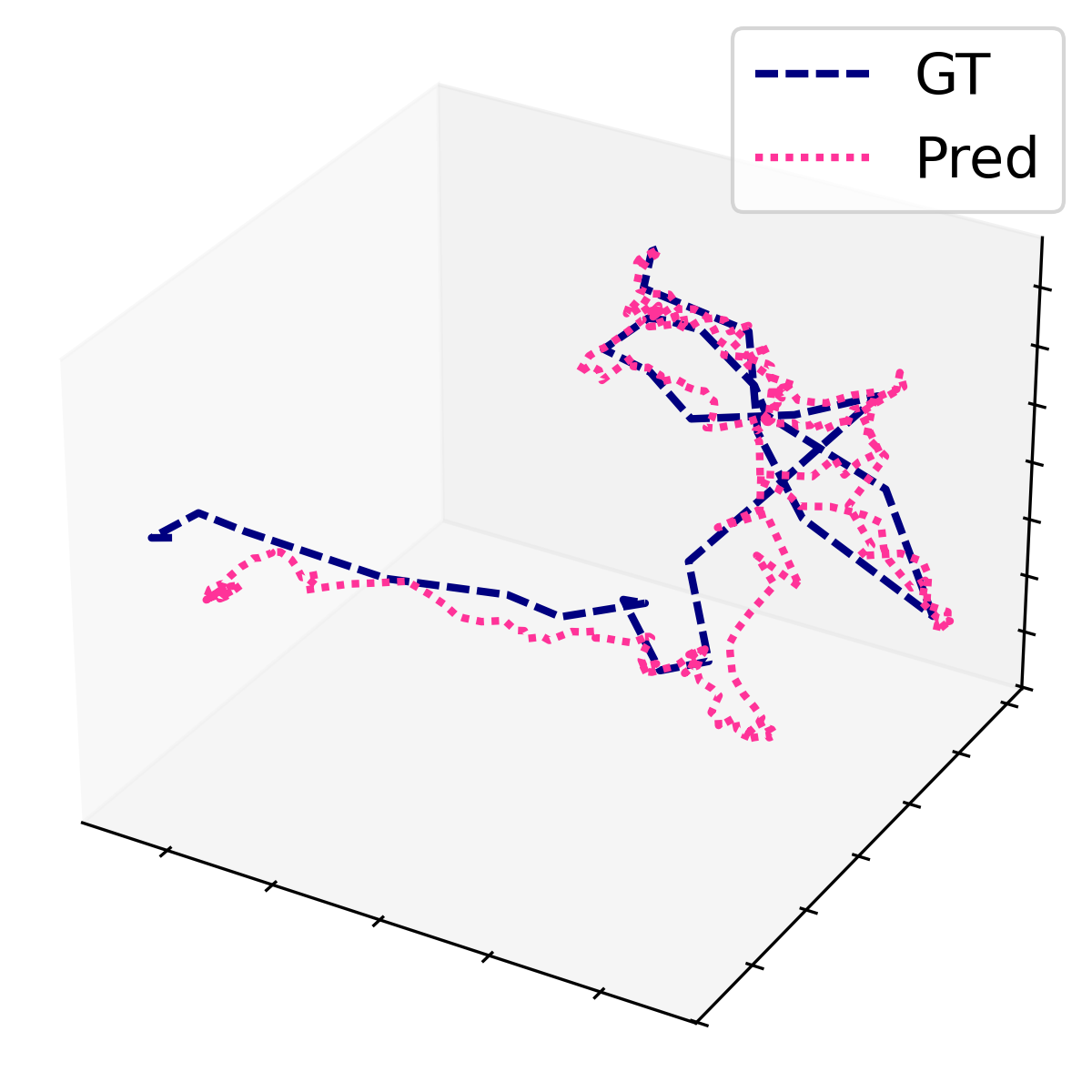} &
            \includegraphics[width=\linewidth, valign=c]
            {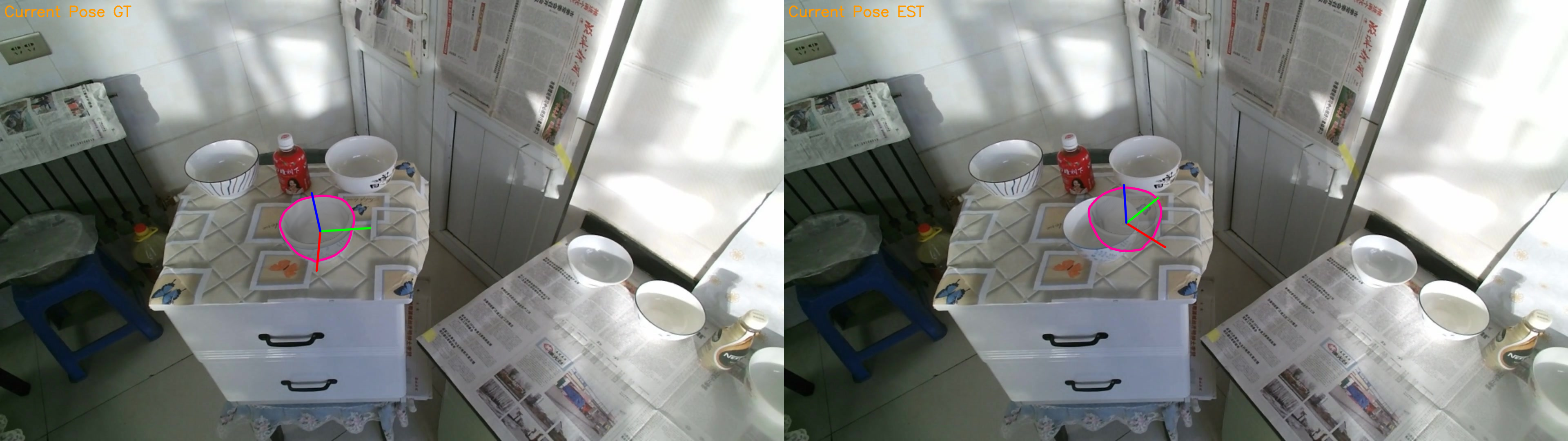}
            \\[2mm]

            \textbf{w. uncertainty} &
            \includegraphics[width=\linewidth, valign=c]
            {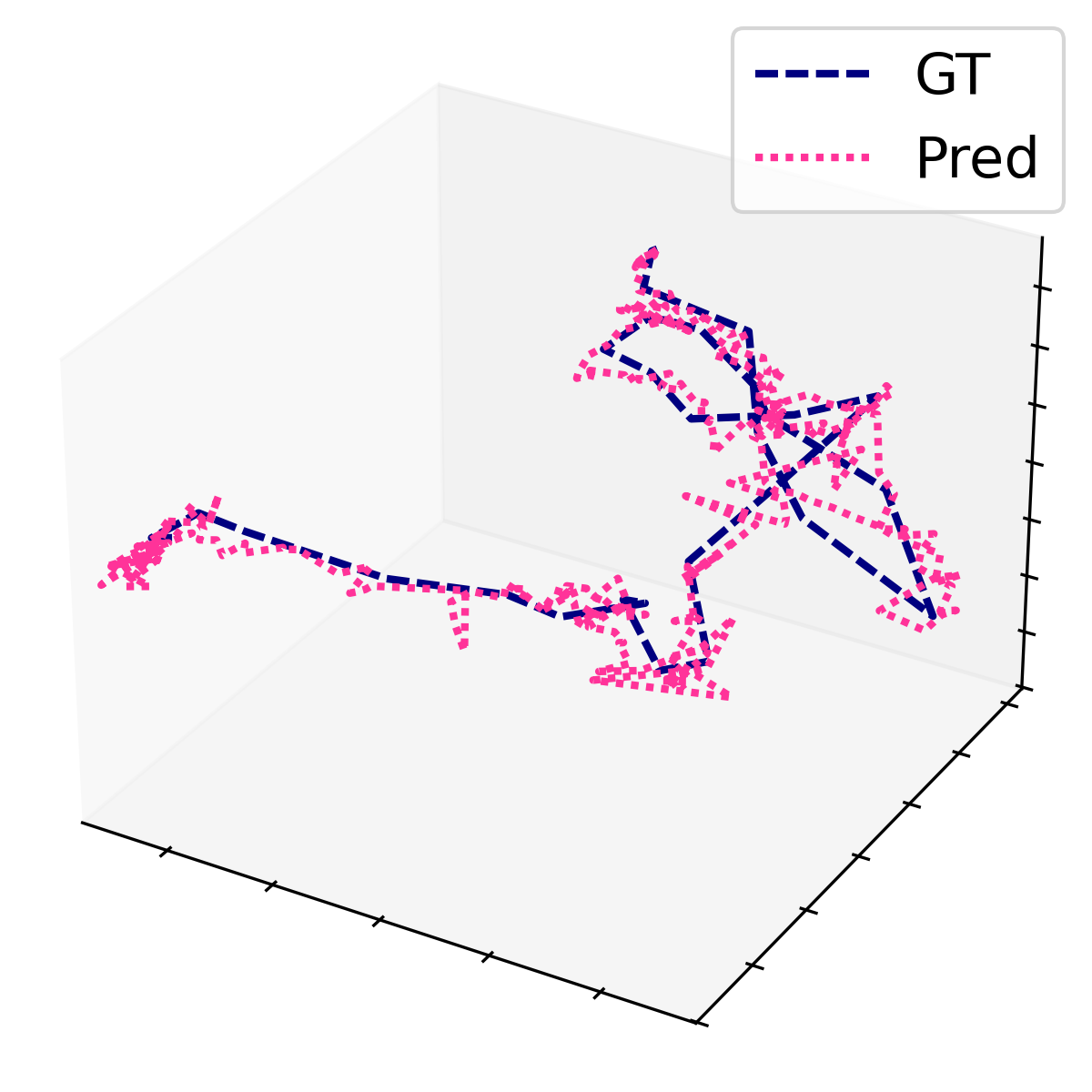} &
            \includegraphics[width=\linewidth, valign=c]
            {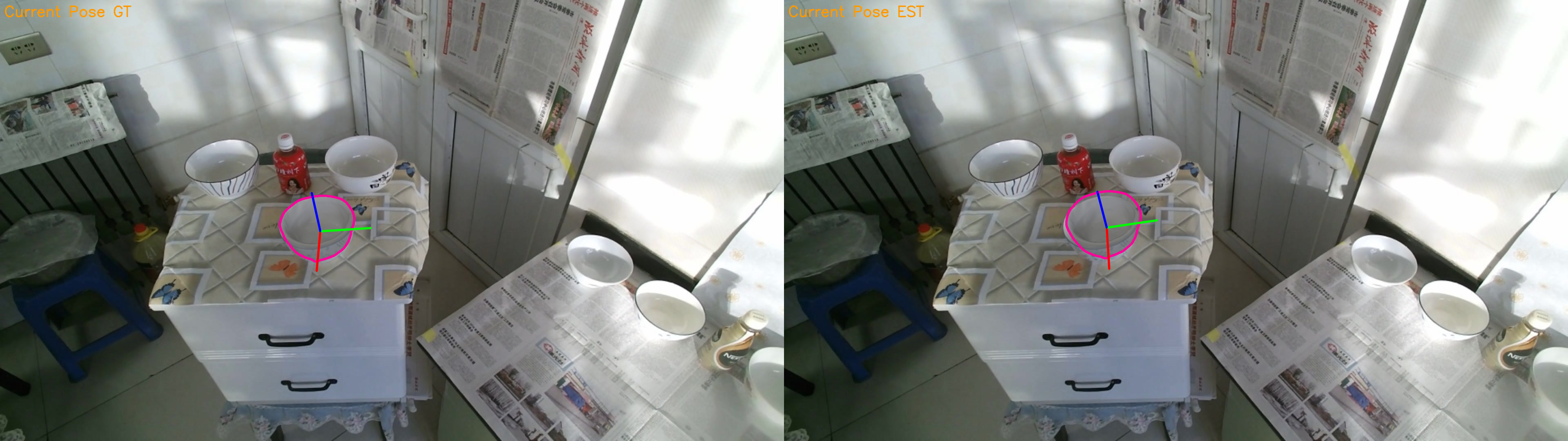}
        \end{tabular}
        \label{fig:weighted_textureless}
    \end{subfigure}

    \vspace{3mm}

    \begin{subfigure}{\linewidth}
        \centering
        \begin{tabular}{
            @{}>{\centering\arraybackslash}m{0.12\linewidth}
               >{\centering\arraybackslash}m{0.20\linewidth}
               >{\centering\arraybackslash}m{0.62\linewidth}@{}
        }
            \textbf{w/o uncertainty} &
            \includegraphics[width=\linewidth, valign=c]
            {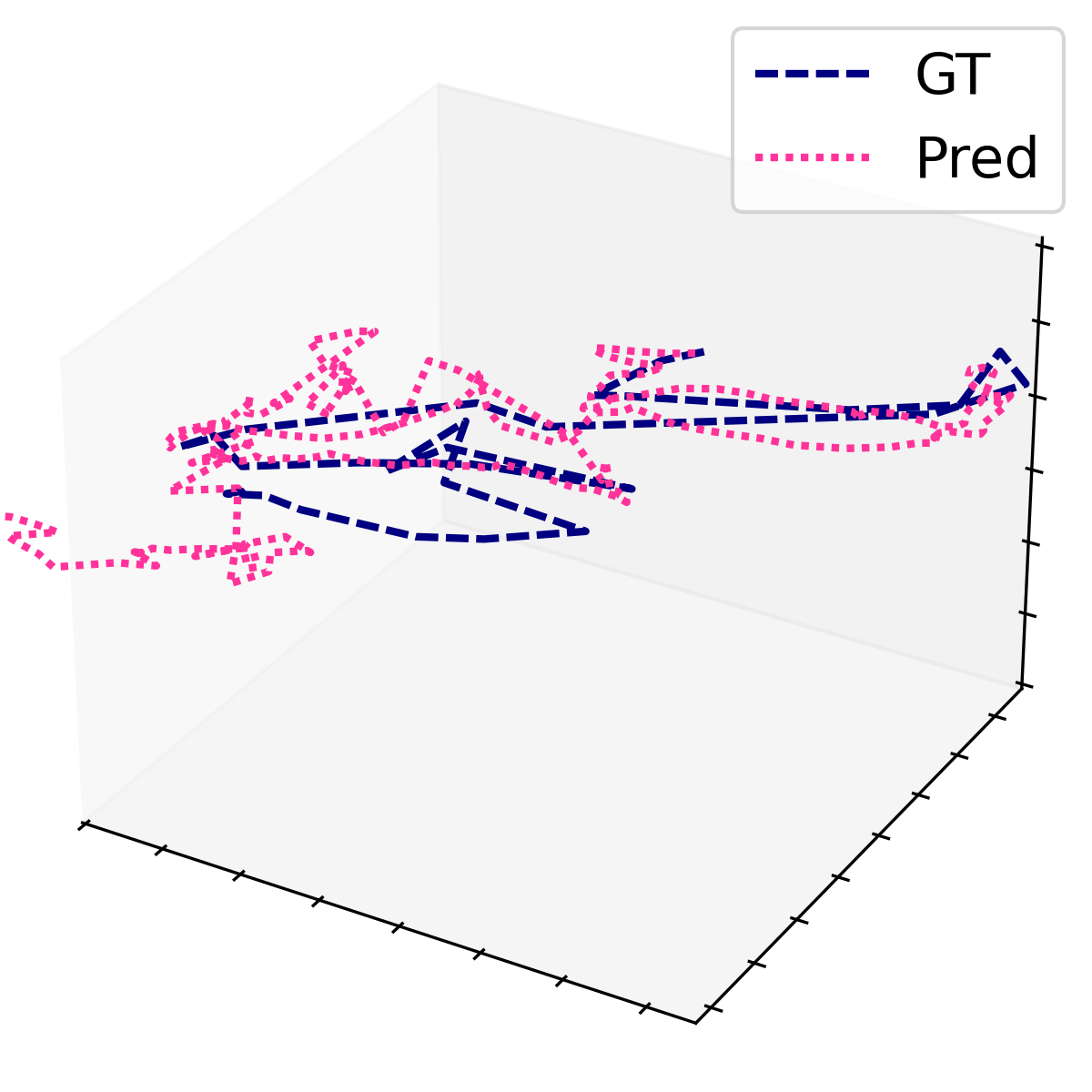} &
            \includegraphics[width=\linewidth, valign=c]
            {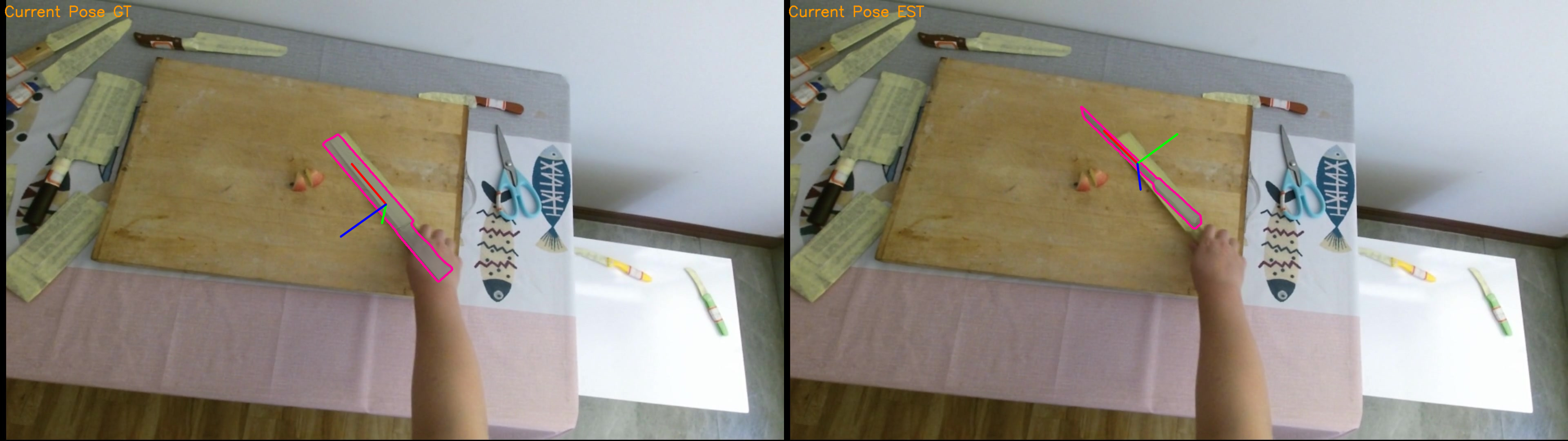}
            \\[2mm]

            \textbf{w. uncertainty} &
            \includegraphics[width=\linewidth, valign=c]
            {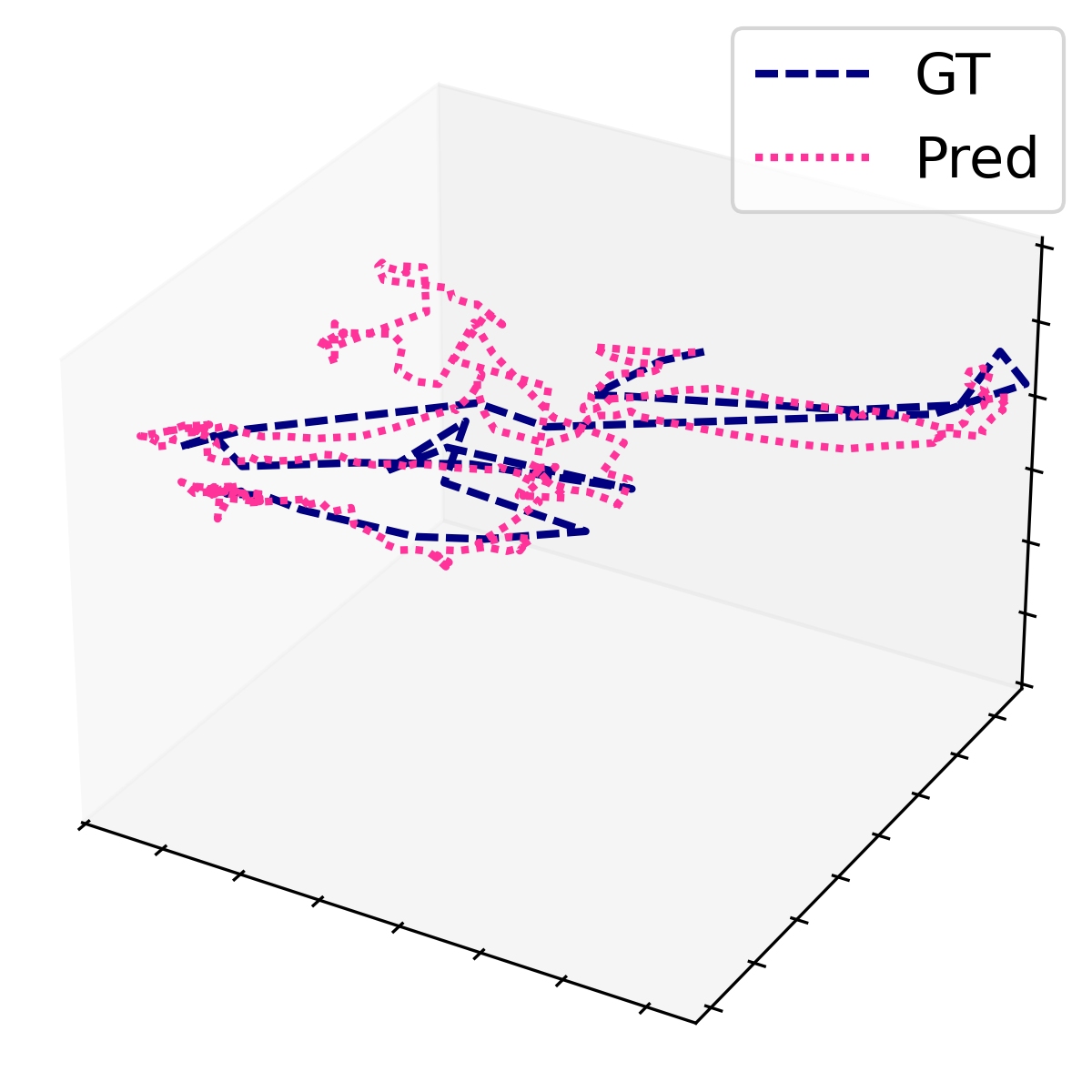} &
            \includegraphics[width=\linewidth, valign=c]
            {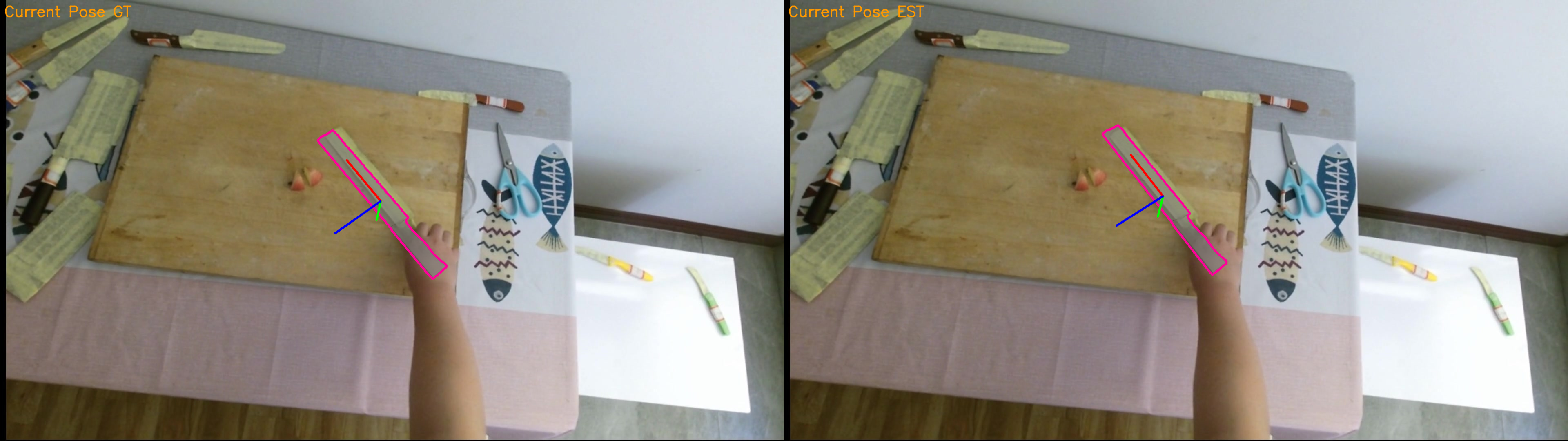}
        \end{tabular}
        
        \label{fig:weighted_thin_1}
    \end{subfigure}

    \vspace{3mm}

    \begin{subfigure}{\linewidth}
        \centering
        \begin{tabular}{
            @{}>{\centering\arraybackslash}m{0.12\linewidth}
               >{\centering\arraybackslash}m{0.20\linewidth}
               >{\centering\arraybackslash}m{0.62\linewidth}@{}
        }
            \textbf{w/o uncertainty} &
            \includegraphics[width=\linewidth, valign=c]
            {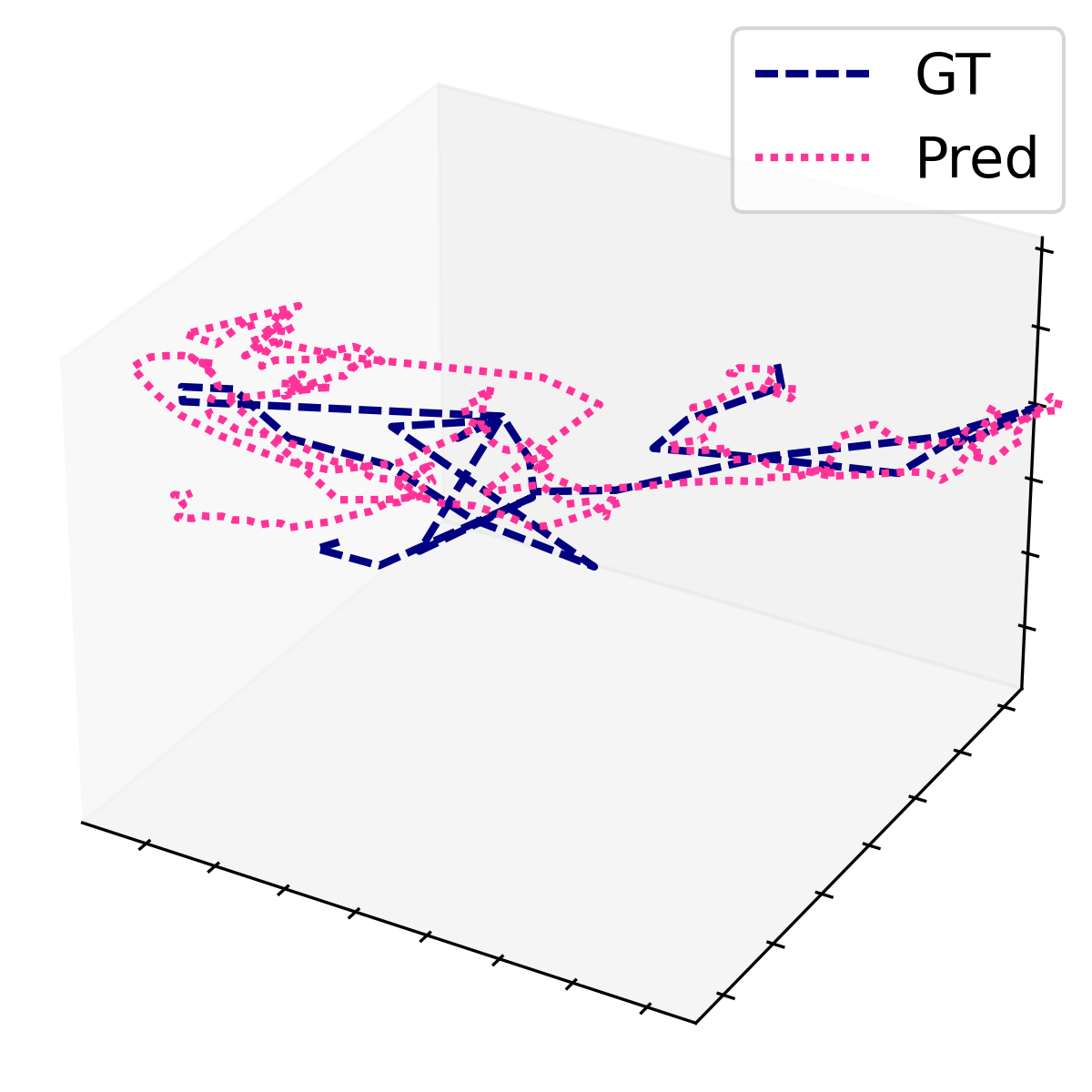} &
            \includegraphics[width=\linewidth, valign=c]
            {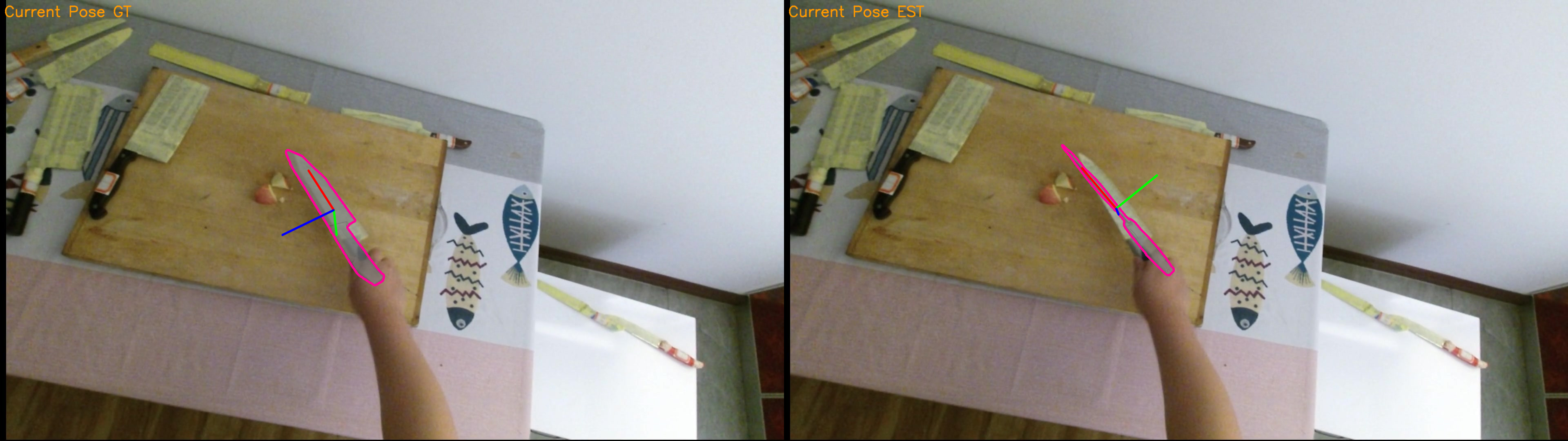}
            \\[2mm]

            \textbf{w. uncertainty} &
            \includegraphics[width=\linewidth, valign=c]
            {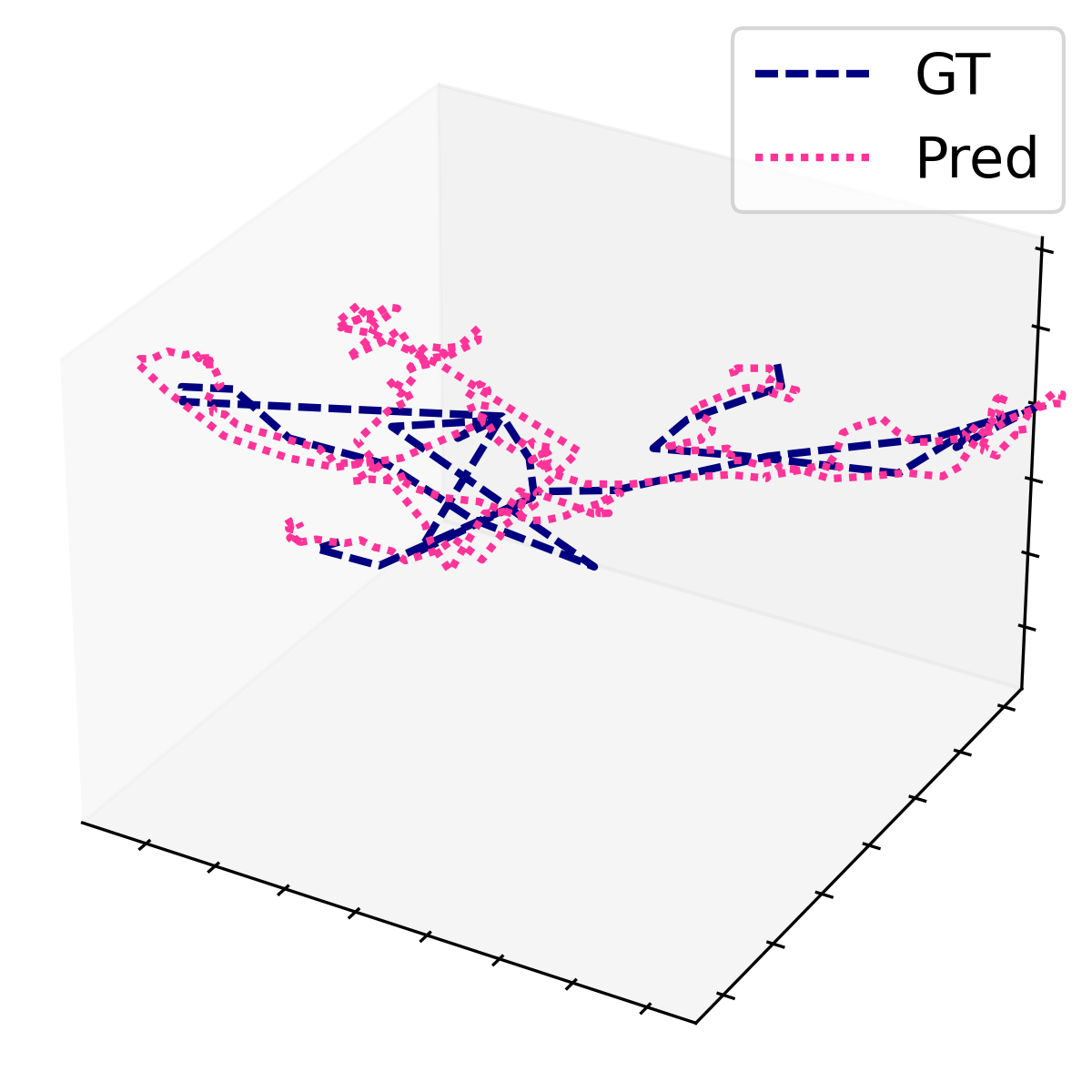} &
            \includegraphics[width=\linewidth, valign=c]
            {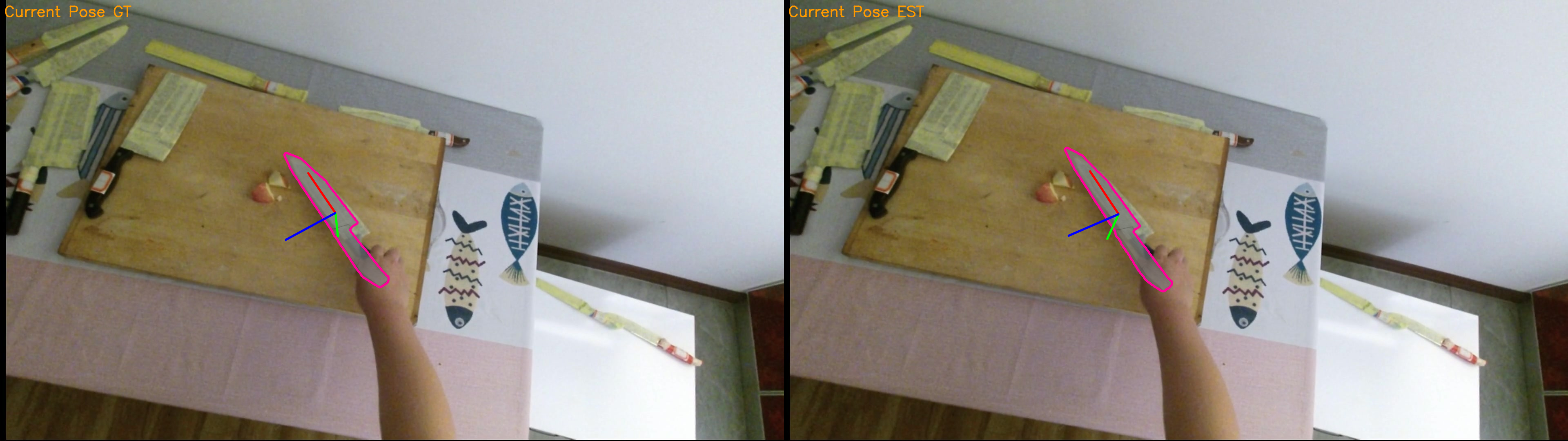}
        \end{tabular}
        
        \label{fig:weighted_thin_2}
    \end{subfigure}

    \caption{
        Qualitative comparison of tracking without and with uncertainty
        modeling. Each pair shows the tracking result without uncertainty
        (\textbf{top}) and with uncertainty (\textbf{bottom}). Modeling uncertainty improves
        robustness in geometrically ambiguous cases, including textureless
        and thin objects.
    }
    \label{fig:weighted_results}
\end{figure*}

\cref{fig:weighted_results} shows the qualitative results of pose tracking using uncertainty-aware PGO compared to unweighted (using identity as the uncertainty) PGO. 

\section{Qualitative Results on Challenge Cases}
\label{sec:challenges}
\begin{figure*}
    \centering

   \begin{tabular}{@{}p{0.3\linewidth}<{\centering}p{0.3\linewidth}<{\centering}p{0.3\linewidth}<{\centering}@{}}
    \footnotesize \textbf{Current Frame} & \footnotesize \textbf{GT Pose} & \footnotesize \textbf{Estimated Pose}
    \end{tabular}
    
    \includegraphics[width=\linewidth]{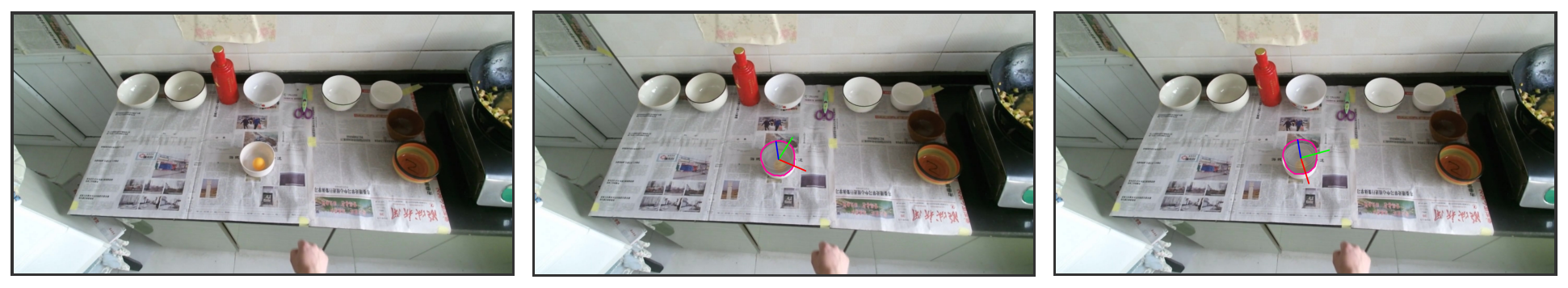}
    \caption{Failure on textureless symmetric object. Poor ADD metric due to the symmetry ambiguity.}
    \label{fig:sym1}
    \centering
    \includegraphics[width=\linewidth]{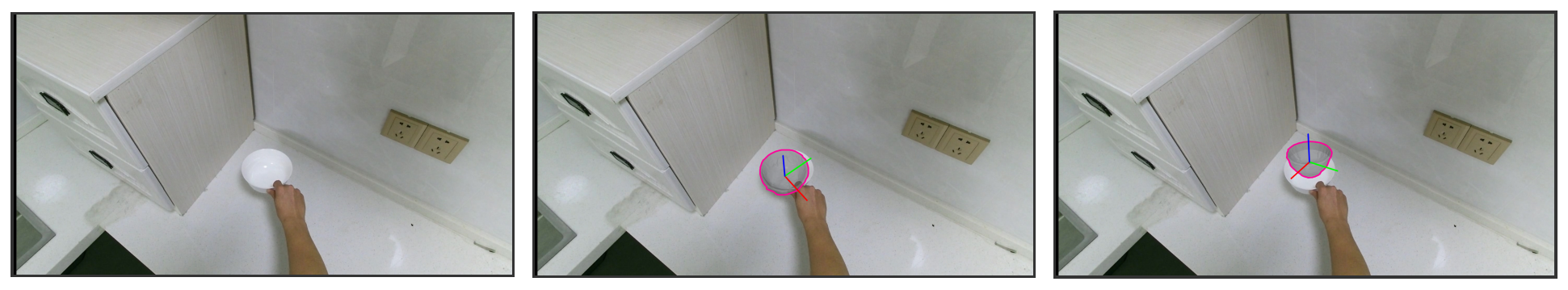}
    \caption{Failure on textureless symmetric object.}
    \label{fig:sym2}
    \centering
    \includegraphics[width=\linewidth]{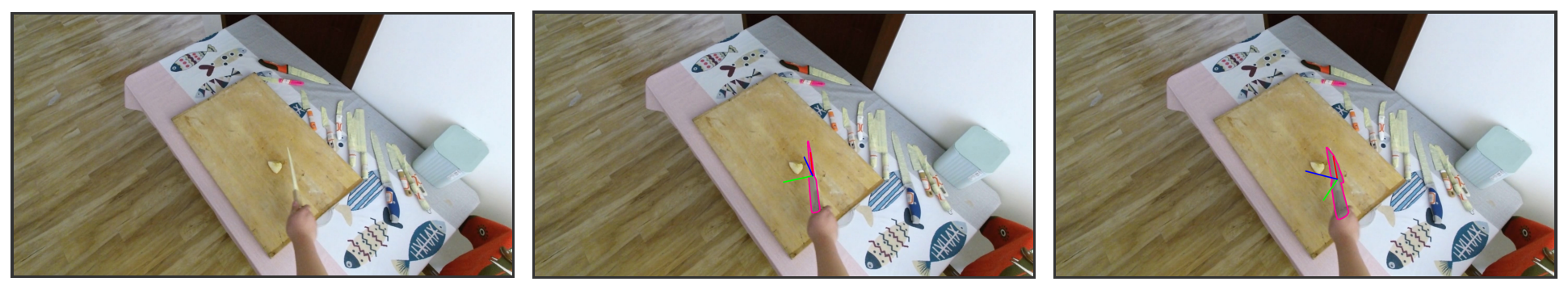}
    \caption{Failure cases when viewing thin long objects from their end-on view axis.}
    \label{fig:thin1}
    \centering
    \includegraphics[width=\linewidth]{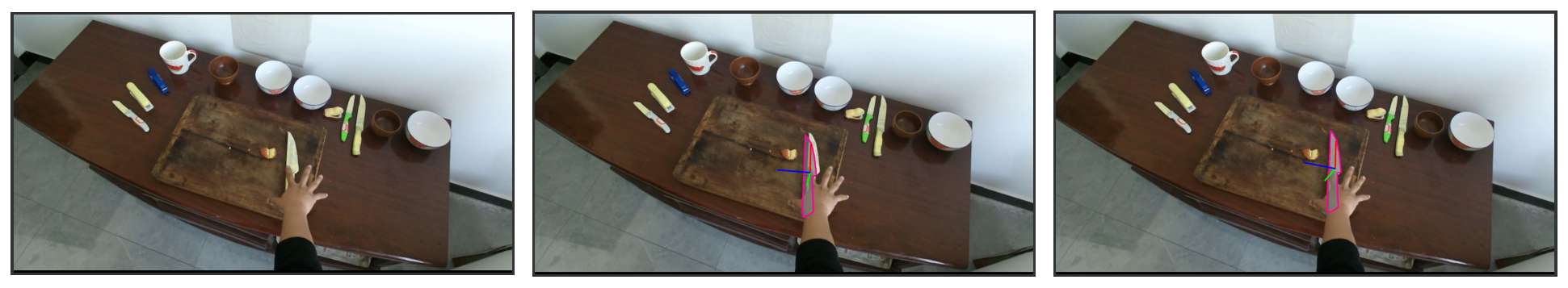}
    \caption{Track thin objects from their non end-on view viewed under heavy hand occlusion.}
    \label{fig:thin2}
    \centering
    \includegraphics[width=\linewidth]{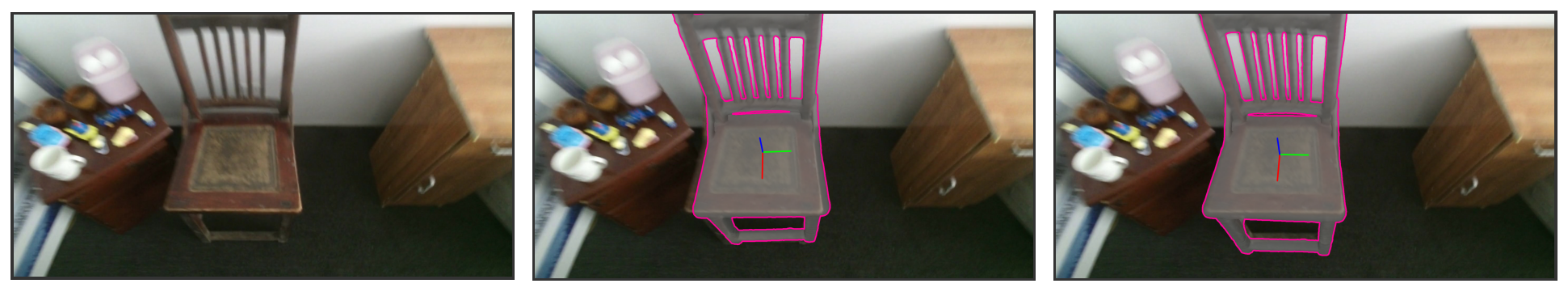}
    \caption{Track the object pose under motion blur.}
    \label{fig:blu1}

\end{figure*}

\begin{figure*}
    \centering
   \begin{tabular}{@{}p{0.3\linewidth}<{\centering}p{0.3\linewidth}<{\centering}p{0.3\linewidth}<{\centering}@{}}
    \footnotesize \textbf{Current Frame} & \footnotesize \textbf{GT Pose} & \footnotesize \textbf{Estimated Pose}
    \end{tabular}
    \includegraphics[width=\linewidth]{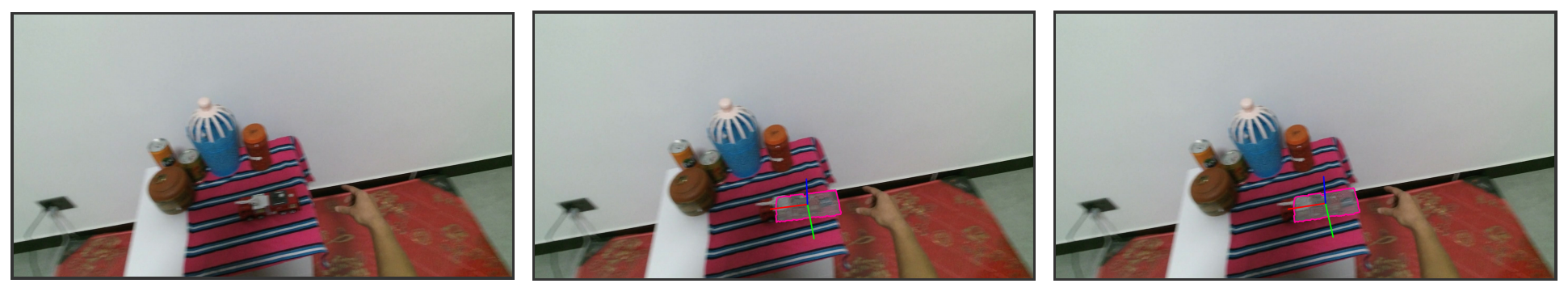}
    \caption{Track the object pose under motion blur.}
    \label{fig:blu2}
    \centering
    \includegraphics[width=\linewidth]{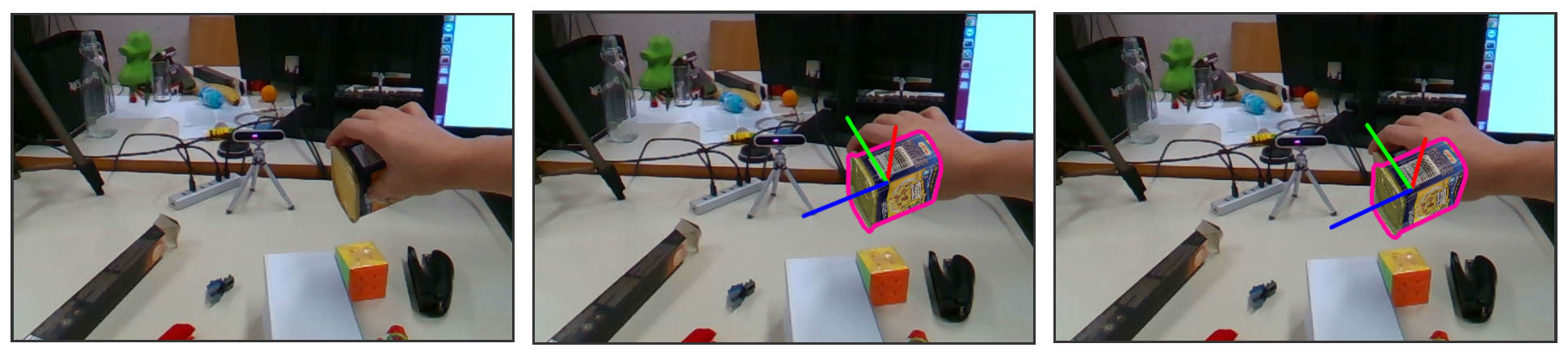}
    \caption{Track the object pose under heavy hand occlusion.}
    \label{fig:occ1}
    \centering
    \includegraphics[width=\linewidth]{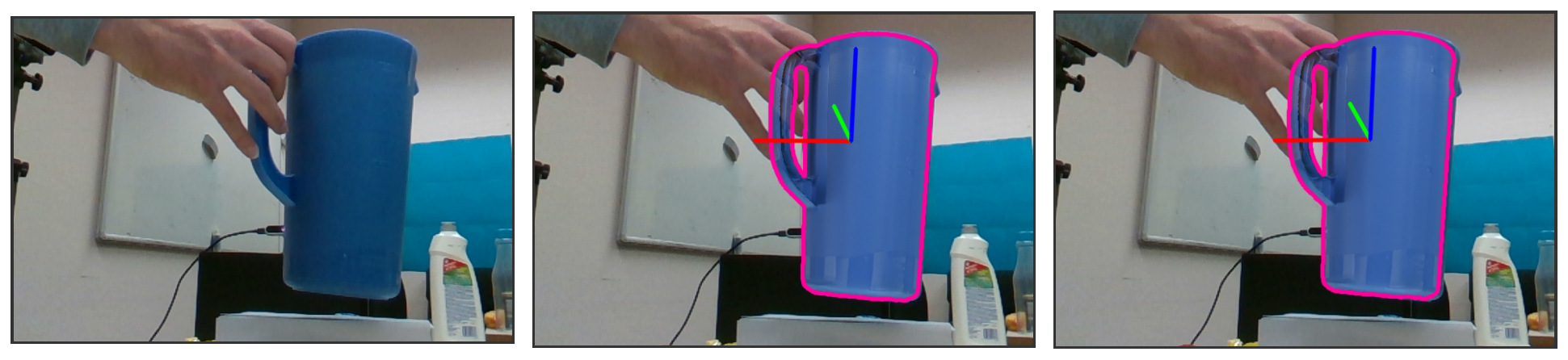}
    \caption{Track the pose of weak texture objects.}
    \label{fig:wt1}
\end{figure*}

\cref{fig:sym1}, \cref{fig:sym2} and \cref{fig:thin1} show the failure cases of our method on textureless symmetric objects and thin objects viewed from their end-on view axis. We also provide other challenges, e.g., thin objects with heavy occlusions (\cref{fig:thin2}), motion blur (\cref{fig:blu1},~\cref{fig:blu2}), heavy occlusion (\cref{fig:occ1}), objects with weak texture (\cref{fig:wt1}), where our method successfully tracks poses.
   

\end{document}